\documentclass[10pt,twocolumn,letterpaper]{article}

\usepackage[pagenumbers]{cvpr} 

\usepackage{tabularx}
\usepackage{multirow} 
\newcolumntype{Y}{>{\centering\arraybackslash}X}
\usepackage{adjustbox}
\usepackage[table]{xcolor}
\usepackage{pifont}
\definecolor{forestGreen}{RGB}{34, 139, 34}
\definecolor{firebrick}{RGB}{178, 34, 34}

\newcommand{\myxmark}{\color{firebrick}{×}}
\newcommand{\mycheckmark}{\color{forestGreen}{\checkmark}}

\definecolor{cvprblue}{rgb}{0.21,0.49,0.74}
\usepackage[pagebackref,breaklinks,colorlinks,allcolors=cvprblue]{hyperref}
\def\paperID{13165} 
\def\confName{CVPR}
\def\confYear{2026}

\title{Q-SAM2: Accurate Quantization for Segment Anything Model 2}

\author{Nicola Farronato\(^{\diamondsuit \clubsuit}\) \hfill Florian Scheidegger\(^{\diamondsuit}\) \hfill Mattia Rigotti\(^{\diamondsuit}\) \vspace{0.25cm}\\
Cristiano Malossi\(^{\diamondsuit}\) \hfill Michele Magno\(^{\clubsuit}\)\hfill Haotong Qin\(^{\clubsuit}\) \vspace{0.25cm} \\
\(\diamondsuit\) IBM Research Zurich \vspace{0.1cm}\\
\(\clubsuit\) ETH Zurich \vspace{0.1cm}\\
{\tt\small nicola.farronato@ibm.com}  \; \;  
{\tt\small \{eid, mrg, acm\}@zurich.ibm.com}\\
{\tt\small \{michele.magno, haotong.qin\}@pbl.ee.ethz.ch
}
}
\begin{document}
\maketitle

\begin{abstract}
The Segment Anything Model 2 (SAM2) is a powerful foundation model for promptable segmentation. However, its high computational and memory costs are a major barrier to deployment on resource-constrained devices. In this paper, we present Q-SAM2, an accurate low-bit quantization method that achieves high compression and high fidelity. To address performance degradation arising from challenging weight and activation distributions during quantization, Q-SAM2 introduces two novel contributions: Variance-Reduced Calibration (VRC), an initialization method that reduces weight statistical variance by minimizing the Frobenius norm over a small calibration batch; and Learnable Statistical Clipping (LSC), a Quantization-Aware Training (QAT) method that learns momentum-stabilized clipping factors to manage outliers in weights and activations. Comprehensive experiments demonstrate that Q-SAM2 achieves highly accurate inference with substantial efficiency gains, significantly surpassing state-of-the-art general QAT schemes, particularly in the ultra-low 2-bit regime. Specifically, Q-SAM2 achieves an accuracy gain of up to 9.7 ppt in J\&F on the video segmentation benchmark and 7.3 ppt in mIoU for instance segmentation over the best competing QAT model, all while achieving an 8x reduction in model size compared to the BF16 baseline.
\end{abstract}

\section{Introduction}
\label{sec:intro}
The Segment Anything Model 2 (SAM2) has emerged as the foundational framework for unified promptable segmentation across static images and video~\cite{ravi2024sam2}.
Its performance on video segmentation makes this model attractive for applications in robotics and tiny machine learning. However, taking advantage of real-time inference with this model requires high-end hardware, such as an NVIDIA A100 GPU. 
Furthermore, edge deployment scenarios typically impose tighter memory and storage constraints, even at the MB level in extreme cases, \cite{han2016deepcompressioncompressingdeep}, which makes the use of the existing version of SAM2 prohibitive.
A general effective strategy to compress and speed up deep neural networks is to quantize them using fewer bits. 
Quantization is a widely adopted technique for optimizing deep learning models, particularly when deploying them on resource-constrained hardware \cite{gholami2022survey,nagel2021white}. By reducing the numerical precision of weights and activations, quantization enables faster inference, lower power, and smaller model sizes \cite{jacob2018quantization,Banner2019}. 

\begin{figure}[t]
    \centering
    \includegraphics[width=\linewidth]{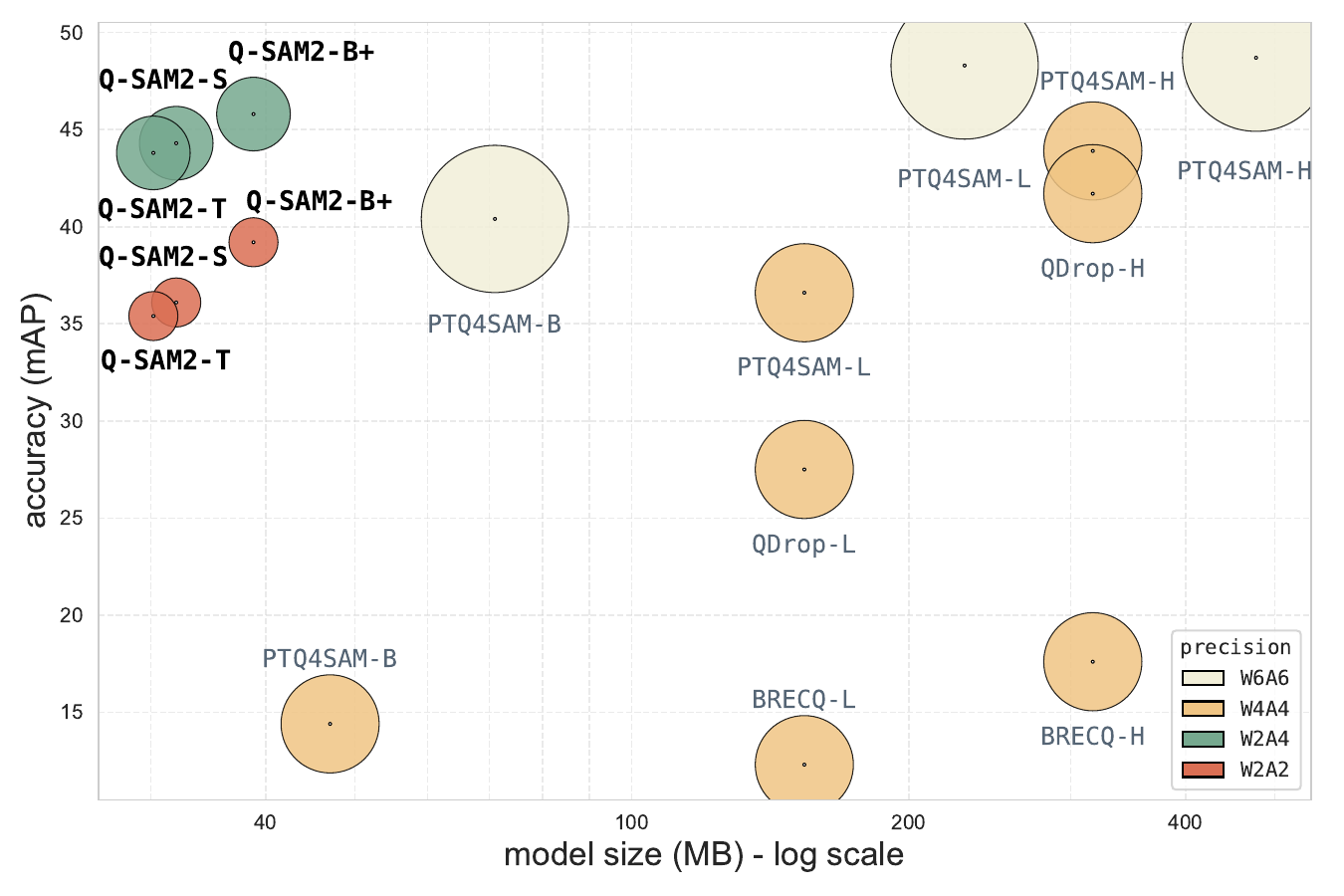}
    \caption{Comparison between Q-SAM2 and Post-Training Quantization (PTQ) based on SAM~\cite{kirillov2023segany} [ViT-B/L/H]. Q-SAM2 defines a new SOTA Pareto frontier; even at its smallest precision (W2A2), it maintains accuracy comparable to higher-precision SOTA methods (PTQ4SAM~\cite{PTQ4SAMLv}, BRECQ~\cite{li2021brecq}, QDROP\cite{wei2023qdroprandomlydroppingquantization}).}
    \label{fig:ptq_comparison}
\end{figure}

\begin{figure*}[t]
    \centering
    \includegraphics[width=\linewidth]{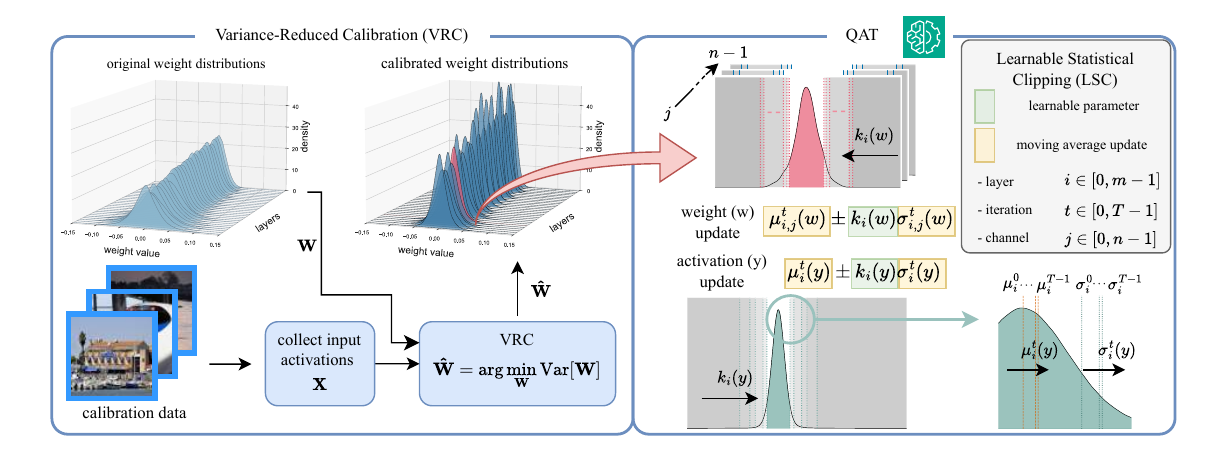}
    \caption{The Q-SAM2 approach. The weight distributions of the linear layers in the image encoder are calibrated using the VRC to reduce variance. We substitute the original encoder and train the network using the LSC in a QAT pipeline.}
    \label{fig:Q-SAM2}
\end{figure*}

For these reasons, the demand for post-training quantization (PTQ) methods has grown substantially in recent years, particularly due to the increasing size and complexity of foundation models like large language models (LLMs)~\cite{Dettmers2022,xiao2023smoothquant,ashkboos2024quarot,liu2025spinquantllmquantizationlearned,frantar-gptq,Nagel2019Data-FreeQuantization,PTQ4SAMLv,wu2024ptq4dit}. These models often consist of billions of parameters, making them difficult to deploy efficiently without compression techniques. PTQ offers a practical solution by enabling model compression without requiring retraining, thus reducing computational costs and accelerating deployment \cite{gholami2022survey, nagel2020adaround,li2021brecq}. Quantization Aware Training (QAT)~\cite{Cai2017,zhou2016dorefa,esser2019lsq,choi2018pact,Hubara2017Quantizedneuralnetworks}, offers a more accurate solution by simulating quantization effects during the training phase, allowing the model to adapt to reduced numerical precision~\cite{jacob2018quantization,wu2020integer,nagel2021white}. QAT enables the model to recover accuracy under aggressive compression, but its adoption remains limited in transformer architectures in resource-constrained scenarios due to its lack of extensive investigation and the high computational cost. 

While no previous work has focused on the SAM2 architecture, recent studies have proposed PTQ algorithms for the original Segment Anything Model \cite{kirillov2023segany}. In particular, \cite{PTQ4SAMLv,PQ-SAMLiu} investigate the quantization challenges of SAM and introduce methods capable of producing per-channel quantized models with few data samples. Although PTQ avoids full retraining, the inability to adapt model parameters during quantization often leads to suboptimal results at ultra-low bit-widths. This limitation is highlighted in Figure \ref{fig:ptq_comparison}, where state-of-the-art PTQ methods already exhibit significant performance degradation at 4-bit precision. The failure of PTQ at 4 bits motivates our work on a more robust, training-aware approach to achieve an accurate 2-bit model.

Quantizing SAM2 is nontrivial, particularly for the image encoder, a hierarchical transformer~\cite{Ryali2023Hiera} which accounts for over $90\%$ of the total parameters (in the B+ model) and poses a significant challenge for ultra-low-bit quantization (see Section~\ref{sec:method}). This difficulty stems from two main issues: the linear layer weight matrices exhibit values that deviate significantly from the distribution center (introducing outliers), and the activation output displays heavy-tailed distributions. Both of these characteristics severely degrade performance when uniform quantization schemes are applied.

To address these challenges and achieve 2-bit precision, we propose Q-SAM2, a complete QAT pipeline. As shown in Figure \ref{fig:Q-SAM2}, our approach introduces two novel components to tackle the critical issues of high weight variance and activation outliers: an initialization method called \textbf{Variance-Reduced Calibration (VRC)}, and a QAT algorithm named \textbf{Learnable Statistical Clipping (LSC)}. This two-step approach first prepares the model for quantization and then robustly trains it, ensuring high accuracy even at 2-bit precision.
\noindent
Our main contributions are as follows:
\begin{itemize}
\item We introduce Q-SAM2, the \textbf{first} quantization pipeline for the SAM2 architecture. Our method establishes a new state-of-the-art for QAT, surpassing the stronger baseline up to 7.3ppt in mIoU. We provide validation of results against SOTA PTQ methods and demonstrate robust accuracy even in ultra-low bit schemes.
\item We propose a novel light calibration method, VRC, that reduces the variance of the encoder's weight distributions using a small batch of images. This procedure significantly lowers the initial quantization error, providing an improved starting point for the subsequent training.
\item We develop a tailored QAT pipeline that leverages the calibrated encoder (from VRC) and a novel algorithm, LSC. The LSC grounds the quantization range using robust EMA-tracked statistical estimates ($\mu, \sigma$). This hybrid design introduces only a single, learnable coefficient ($k$) per quantizer, enabling stable ultra-low-bit convergence.
\end{itemize}
\noindent The paper is organized as follows:
Section~\ref{sec:background_related} describes the preliminaries and related work, Section~\ref{sec:method} shows the Q-SAM2 pipeline, Section~\ref{sec:experiments} presents our experimental setup and results, and Section~\ref{sec:conclusions} concludes our findings.

\section{Preliminaries and Related Work}
\label{sec:relwork}
\label{sec:background_related}
\subsection{Model Quantization}
Quantization is the process of approximating a continuous or high-resolution discrete domain by a finite set of representative levels, to optimize computational efficiency and memory usage with controlled performance degradation. The quantization process can be formulated as:
\begin{equation}
    \mathbf{X}_q = \mathrm{clip}\left( \mathrm{round}\left( \frac{\mathbf{X}}{s} \right) + z,\ 0,\ 2^b - 1 \right).
    \label{eq:quantization}
\end{equation}
Where $\mathbf{X}$ is a high-precision input tensor (e.g., \texttt{FP32}). The process maps $\mathbf{X}$ to the low-bit integer tensor $\mathbf{X}_q$ using the scaling factor $s$ and the integer zero-point $z$. Specifically, $s$ defines the step size of the quantization levels, while $z$ acts as an offset ensuring that the real-valued zero is mapped exactly to an integer. The bit-width $b$ determines the cardinality of the discrete set, typically $[0, 2^b-1]$ for unsigned quantization. The clipping function is mathematically necessary to constrain the projected values $\mathrm{round}(\mathbf{X}/s) + z$ to this finite integer domain. These parameters can be defined as scalars ($s,z \in \mathbb{R}$) for per-tensor quantization, or vectors ($\mathbf{s}, \mathbf{z} \in \mathbb{R}^d$) for per-channel quantization or a tensor (\( \mathbf{s},\mathbf{z} \in \mathbb{R}^{d_1 \times d_2 \times \dots} \)) for per-group quantization. We focus on per-channel uniform quantization, which offers a good balance between efficiency and complexity.

\subsection{Post-Training Quantization (PTQ)}
Post-training quantization (PTQ) enables efficient deployment of large pretrained models by applying quantization without retraining the full network. Scale \(s\) and zero-point \(z\) are typically computed from calibration data. To improve accuracy, various PTQ methods introduce transformations to reduce quantization error: AdaRound~\cite{nagel2020adaround} and BrecQ~\cite{li2021brecq} optimize rounding and reconstruction, respectively. QDROP~\cite{wei2023qdroprandomlydroppingquantization} further refined this by adding a Dropout-inspired regularization, making weight optimization more robust to activation quantization errors. Other approaches leverage Hessian-based and calibration-free formulations like GPTQ~\cite{frantar-gptq} and HQQ~\cite{badri2023hqq}, or minimizing weight outliers via \(L_\infty\)-based regularization as in MagR~\cite{zhang2024magr}. 
Activation outliers, common in transformers~\cite{Dettmers2022,wu2024ptq4dit}, further complicate quantization. Techniques such as SmoothQuant~\cite{xiao2023smoothquant}, SpinQuant~\cite{liu2025spinquantllmquantizationlearned}, and QuaRot~\cite{ashkboos2024quarot} mitigate this by shifting or rotating activations to promote accurate PTQ.

In the vision domain, PTQ4ViT~\cite{Yuan2022PTQ4ViT} tackles non‑Gaussian activations in vision transformers. PTQ4DiT~\cite{wu2024ptq4dit} introduces channel‑wise salience balancing to mitigate salient-channel and temporal activation outliers in Diffusion Transformers. PTQ4SAM~\cite{PTQ4SAMLv} is the first PTQ method tailored to SAM, utilizing bimodal integration and adaptive granularity quantization to transform challenging activation distributions. 

\subsection{Quantization Aware Training (QAT)}
In Quantization Aware Training (QAT), the quantization process is simulated during training through Equation~\ref{eq:quantization}, allowing the network to adapt to quantization noise and minimize accuracy degradation. The scale \( s \) and zero point \( z \) can be treated as learnable parameters or are dynamically adapted, and the model weights themselves are updated through backpropagation to account for quantization effects. This joint adaptation allows the model to recover from quantization-induced distortions and is essential to preserve accuracy under low-bit regimes. Formally, QAT introduces non-differentiable quantization into the training loop, typically handled using straight-through estimators (STE)~\cite{Bengio2013} to enable gradient-based optimization.

Several QAT methods have been proposed to improve accuracy in low-precision regimes. One of the first and most influential works is the MinMax-based QAT approach~\cite{jacob2018quantization}, which uses affine quantization and simulated integer operations during training, enabling efficient 8-bit deployment with integer-only inference. Another effective method is DoReFa-Net~\cite{zhou2016dorefa}, which introduces quantization of weights, activations, and gradients using deterministic functions. However, its use in modern transformer architectures is limited due to the inability to quantize models with large activation ranges \cite{li2022mqbenchreproducibledeployablemodel}. 
PACT~\cite{choi2018pact} introduced a learnable clipping parameter for activations, enabling better control of dynamic range. Nevertheless, this introduces additional parameters and gradients, complicating the deployment. LSQ~\cite{esser2019lsq} is widely regarded as one of the most robust QAT methods, learning the quantization step size via backpropagation. Building on this, LSQ+~\cite{bhalgat2020lsqimprovinglowbitquantization} improves training stability by introducing a log-based gradient formulation for the learnable scale, which prevents the scale from becoming zero and stabilizes the training process.

QAT remains underexplored in transformer-based vision architectures, where challenges such as weight oscillation, activation outliers, and attention-specific sensitivity can severely impact low-bit performance. Existing studies~\cite{Liu2023Oscillation, huang2024quantization} focus solely on LSQ, without evaluating alternative strategies or providing broader comparisons between quantization methods.

\section{Method}
\label{sec:method}
\subsection{Challenges in Quantizing SAM2}
\label{sec:challenges}
The SAM2 architecture is built on a Hiera-based encoder~\cite{Ryali2023Hiera}, which hierarchically merges tokens for an efficient multiscale representation. While effective for dense vision tasks, this architecture introduces challenges for quantization due to the presence of outliers in both weights and activations. We find that weight distributions are heavy-tailed, with extreme values concentrated in early layers, leading to instability under low-bit quantization. Activation outliers, though sparse, consistently appear in specific channels, further degrading precision. Together, these factors cause significant performance drops when reducing weight precision to 4 or even 2 bits.

As a preliminary investigation, we analyze explicit value clipping on weight matrices to mitigate outliers by truncating extreme values to increase resolution in dense regions. While this results in an initial accuracy drop, the preservation of core predictive behavior indicates that these outliers are non-essential for effective functioning. We observe the same behavior for activations, which are even more difficult to quantize on low bits because of their higher value range. The complete findings and supporting analysis for this preliminary investigation are detailed in Appendix~\ref{sec:distributions}. 
Our method is built on a key insight: since outliers can be clipped without destroying the model's predictive power, the problem becomes optimizing the remaining, dense part of the distribution. We therefore propose a two-stage pipeline: first, we apply \textbf{Variance-Reduced Calibration (VRC)} (Section \ref{sec:VRC}), a novel initialization that narrows the encoder's weight distributions. This provides a low variance, so a lower quantization error starting point. Second, we apply \textbf{Learnable Statistical Clipping (LSC)} (Section \ref{sec:LSC}), a robust QAT algorithm that learns optimal clipping thresholds for both calibrated weights and activations. 

\subsection{Variance-Reduced Calibration (VRC)}
\label{sec:VRC}

\begin{figure*}[htb]
  \centering
    \includegraphics[width=\textwidth]{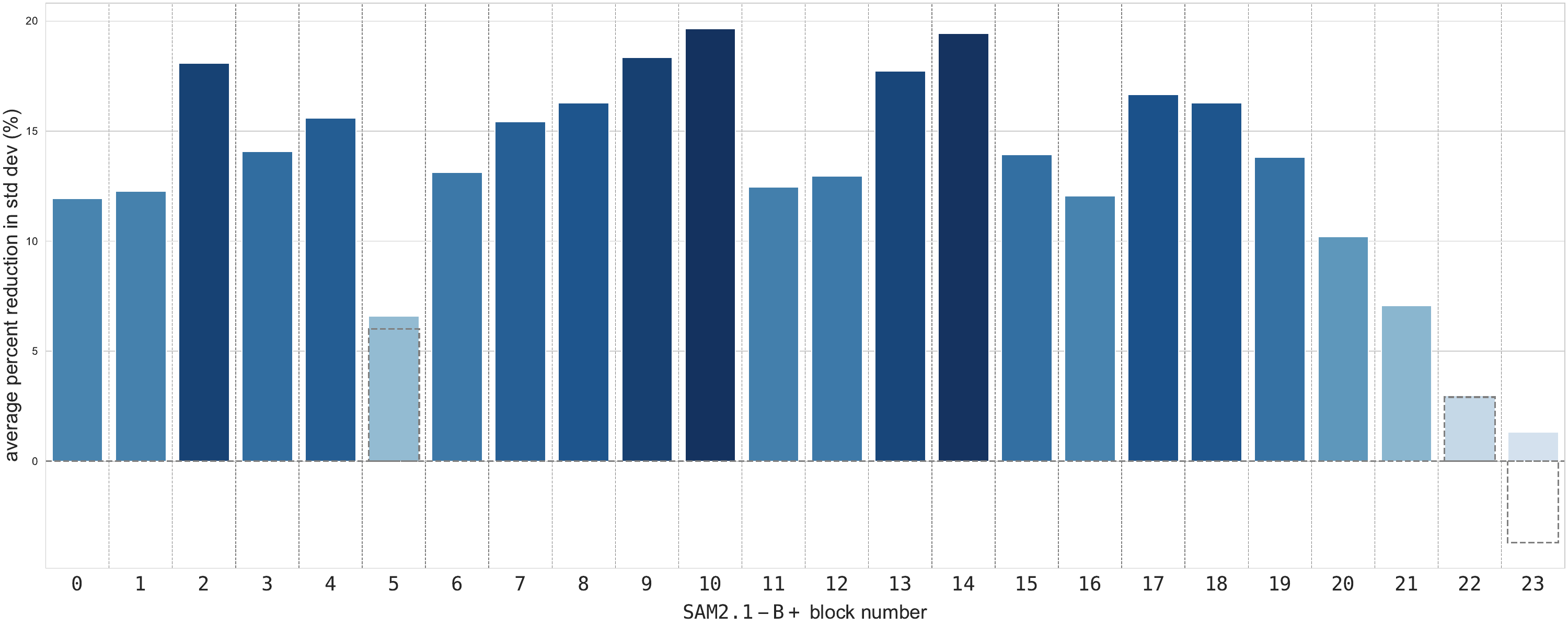}
  \caption{Impact of VRC on SAM2.1-B+ image encoder weight distributions for $\lambda_0=2.0$. VRC achieves a significant (10-20\%) reduction in standard deviation, averaged per transformer block. This compresses the dynamic range, critically lowering the initial quantization error for the subsequent QAT.}
  \label{fig:VRC_lambda2}
\end{figure*}

Although weight calibration is common in PTQ, the idea of introducing a dedicated preprocessing step before QAT has received little attention~\cite{li2022mqbenchreproducibledeployablemodel}. In this work, we show that VRC can improve performance, particularly in ultra-low bit-width regimes.

Designed as a precursor to our clipping-based QAT, our calibration minimizes weight variance to compact the distribution. This allows for a smaller quantization step size, directly reducing the quantization error for the majority of the weights. We employ a computationally efficient, closed-form minimization of the Frobenius norm using a small calibration set. This approach is theoretically grounded: since the weight means are generally close to zero, the Frobenius norm serves as an accurate proxy for variance.

However, any modification of a fully trained linear layer weight matrix risks degrading the accuracy of the model.
A straightforward but crucial observation is that the accuracy degradation in the final model is directly related to the discrepancy between the output of the linear layers before and after quantization, rather than the quantization error of the matrix elements themselves. 
Any changes of $\mathbf{W}$ will therefore need to balance the compression gains due to quantization with the resulting errors at the output of linear layers.

Based on this insight, we propose a quantization procedure aimed at optimizing this trade-off through an initial calibration step that projects the trained weights onto the subspace spanned by a calibration batch of data samples, which ensures that the layer's outputs are preserved on these known inputs.
In particular, we propose to perform this projection learning the \emph{Moore-Penrose Pseudoinverse} between the input activations and the output of the linear layer while running a forward pass of the original trained model on the calibration samples.
The use of the Moore-Penrose pseudoinverse is motivated by its known properties in terms of 1) minimizing the residual with the calibration output activations and 2) minimizing the Frobenius norm of the obtained weights.
In other words, this simultaneously achieves that the dynamic range of the weight distribution is minimized, making them easier to quantize, while at the same time minimizing the change of output of the linear layer (at least on the calibration samples).

While the residual between pre- and post-calibration output activations is guaranteed to be minimized only on the calibration samples, the Moore-Penrose Pseudoinverse has additional properties that make it suitable as a preprocessing step before quantization.
The fact that it allows us to control the singular values of the calibrated weight matrix (see later the discussion on Tikhonov regularization) implies that the calibration process ``cleans'' its spectrum by eliminating components that are irrelevant, unstable, or poorly supported by the calibration data, leading to a simpler and more robust solution. As a result, the modified layer maintains essential functionality on the calibration inputs while promoting smoother and potentially more general behavior elsewhere, which is particularly important for robustness under quantization or other perturbations.

The selection of calibration data is critical: larger batches improve faithfulness to the original model, while smaller ones reduce memory and runtime costs. 
Moreover, choosing representative frames is equally important because stacking ill-conditioned inputs in the calibration set can collapse the key directions learned during training, harming generalization.
As a result, the layer may perform well on calibration samples, but fail on unseen or diverse data.
Henceforth, calibration inputs must be diverse, well-conditioned, and representative of the operational domain of the model.

The condition number of activations reflects the calibration input quality: low values indicate stable and rich subspaces. However, aggregating inputs presents a risk: a single ill-conditioned sample can propagate instability to the entire batch, compromising the overall projection quality. Since pre-filtering inputs is computationally infeasible, we employ regularization to suppress ill-conditioning, enabling random input selection.

A linear layer is defined by \( f(\mathbf{X}, \mathbf{W}) \).
It maps an input tensor \( \mathbf{X} \in \mathbb{R}^{n \times d_{\text{in}}} \), where \( n \) is the batch size and \( d_{\text{in}} \) is the input feature dimension, to an output tensor \( \mathbf{Y} \in \mathbb{R}^{n \times d_{\text{out}}} \), where \( d_{\text{out}} \) is the output feature dimension, through an affine transformation defined as \( f(\mathbf{X}, \mathbf{W}) = \mathbf{X} \mathbf{W}^\top + \mathbf{b} \), with \( \mathbf{W} \in \mathbb{R}^{d_{\text{out}} \times d_{\text{in}}} \) and \( \mathbf{b} \in \mathbb{R}^{d_{\text{out}}} \).
We consider \(\mathbf{b}=\mathbf{0}\) for the following analysis, without loss of generality.

We compute the minimum Frobenius norm weight matrix \(\mathbf{\hat{W}}\) for the batch \(\mathbf{X}\) using the Moore-Penrose pseudoinverse \cite{Penrose_1955}.
For \( \mathbf{X} \in \mathbb{R}^{n \times d} \) and \( \mathbf{Y} \in \mathbb{R}^{n \times p} \), among all matrices \( \mathbf{W} \in \mathbb{R}^{p \times d} \) satisfying
\(
\mathbf{Y} = \mathbf{X} \mathbf{W}^\top,
\)
 the unique matrix minimizing the Frobenius norm \( \| \mathbf{W} \|_F \) is
\begin{equation}
\label{eq:pseudoinv}
\mathbf{\hat{W}}^\top = \mathbf{X}^\dagger \mathbf{Y},
\end{equation}
where \( \mathbf{X}^\dagger \) denotes the Moore--Penrose pseudoinverse of \( \mathbf{X} \).

To mitigate the instability caused by poorly conditioned calibration samples, we applied the Tikhonov regularization during the projection. From Equation~\eqref{eq:pseudoinv} the new formulation is derived:
\begin{equation}
    \mathbf{\hat{W}}^\top = (\mathbf{X}^\top\mathbf{X}+\lambda I)^{-1} \mathbf{X}^\top\mathbf{Y}
\label{eq:regularization}
\end{equation}
where \(I\) is the $d\times d$ identity matrix, and \(\lambda \in \mathbb{R}^+\) is the cutoff hyperparameter. 
We set \(\lambda\) based on the singular values of the activation matrix to ensure stability during pseudoinverse computation. When the matrix is ill-conditioned but not rank-deficient, \(\lambda\) is chosen as a small multiple of the smallest non-zero singular value. In cases where the smallest singular value indicates an effective rank deficiency (e.g., in low-precision formats like \texttt{bfloat16}), we instead select the smallest stable singular value and scale it accordingly. In practice, \(\lambda=\lambda_0 \sigma_*\),  where $\sigma_*$ is the smallest singular value that avoids rank deficiency, and $\lambda_0$ ranges between one and five. In Appendix~\ref{sec:calibration} we present an analysis of our calibration method and the role of \( \lambda \) in Equation~\ref{eq:regularization}.
Figure~\ref{fig:VRC_lambda2} quantifies the efficacy of our VRC method on the SAM2.1-B+ model (using $\lambda_0 =2.0$). The chart plots the average percent reduction in standard deviation across the linear layers within each transformer block. The dotted blocks contain layers with negative gains due to high instability in the input activations, in which the selected $\lambda_0$ is not sufficient to reduce the variance. In this case, the algorithm applies a fallback to the original layer, resulting in the normal colored block. Most blocks achieve a significant reduction of more than 10\%, with some approaching 20\%. By systematically reducing the weight variance, the VRC compresses the dynamic range of the distributions. This reduction in the initial quantization error is critical for providing a stable and low-error starting point for the subsequent QAT phase. We report the calibrated encoder layer-by-layer in Appendix~\ref{sec:calibration}.

\subsection{Learnable Statistical Clipping (LSC)}
\label{sec:LSC}

Although learnable QAT methods like LSQ \cite{esser2019lsq} and LSQ+ \cite{bhalgat2020lsqimprovinglowbitquantization} are effective, their ``free-floating'' learnable parameters (e.g., step-size $s$ and offset $z$) are not statistically grounded. This is a known source of instability, particularly in the per-channel, low-bit regime, where the large number of sensitive parameters leads to instability or variance, requiring a great deal of hyper-parameter tuning and initialization \cite{bhalgat2020lsqimprovinglowbitquantization}.

To solve this, we introduce \textbf{Learnable Statistical Clipping (LSC)}, a hybrid method that replaces these unstable learnable parameters with robust, statistically-grounded estimates. LSC anchors the quantization bounds to the data's true distribution, leaving only a single, well-behaved parameter to be optimized for each quantizer.

First, we track the first two moments of the distribution using an Exponential Moving Average (EMA), analogous to the robust estimators in Batch Normalization \cite{ioffe2015batchnormalizationacceleratingdeep}. For weights, we track, on each layer, the per-channel batch statistics $\hat{\mu}_{w,c}^{(t)}$ and $\hat{\sigma}_{w,c}^{(t)}$ at iteration $t$:
\begin{align*}
\mu_{w,c}^{(t)} &= \beta\, \mu_{w,c}^{(t-1)} + (1-\beta)\, \hat{\mu}_{w,c}^{(t)},  \\
\sigma_{w,c}^{(t)} &= \beta\, \sigma_{w,c}^{(t-1)} + (1-\beta)\, \hat{\sigma}_{w,c}^{(t)}, 
\end{align*}
where $\beta \in [0,1)$ is the momentum term. We apply the same statistical tracking principle at the per-tensor level for activations.

The impact of this temporal smoothing is different for the two components. Activation batch statistics ($\hat{\mu}_{a}^{(t)}, \hat{\sigma}_{a}^{(t)}$) are inherently noisy, as they are highly dependent on the input data. The momentum $\beta$ is therefore essential for providing a stable, time-averaged estimate of the activation distribution. In contrast, weight statistics are much more static, especially as training converges. While the EMA still provides stabilization, its primary role for weights is to establish a robust estimate during the initial phase of QAT.

Second, we introduce a single, learnable, tensor-wide parameter $k$ for each quantizer. This learnable $k$ is optimized via backpropagation to find the optimal clipping range. This allows the network to learn the best balance between clipping error (the information lost from truncating outliers, which dominates when $k$ is small) and quantization error (the loss of precision from using wide buckets, which dominates when $k$ is large). 

The final dynamic ranges of the quantized values for weights, as presented in Equation~\ref{eq:quantization}, are then defined symmetrically around the tracked mean:
\begin{align*}
\text{min}_{w,c}^{(t)} &= \mu_{w,c}^{(t)} - |k_w|\, \sigma_{w,c}^{(t)}, \\
\text{max}_{w,c}^{(t)} &= \mu_{w,c}^{(t)} + |k_w|\, \sigma_{w,c}^{(t)}. 
\end{align*}
The per-tensor activation bounds ($\text{min}_{a}^{(t)}, \text{max}_{a}^{(t)}$) are computed analogously using their respective per-tensor statistics and learnable coefficient $k_a$.

This formulation is fundamentally more stable. A single layer in LSQ+ requires optimizing thousands of sensitive, independent per-channel parameters ($s_c, z_c$) for weights alone. In contrast, our LSC replaces this entire set with robust, non-learnable statistics ($\mu_{w,c}^{(t)}, \sigma_{w,c}^{(t)}$) and just one learnable scalar, $k_w$. The optimization landscape is thus dramatically simplified. The gradients for $k_w$ and $k_a$ are well-behaved sums aggregated across all channels or tensor elements and are mediated by the stable $\sigma^{(t)}$ statistics.

\section{Experiments}
\label{sec:experiments}
\begin{table*}
\tiny
\centering
\setlength{\tabcolsep}{2mm}
\resizebox{ \linewidth}{!}{
\begin{tabular}{lcl|c|c|c|c|c|c|c|c|c}
\toprule
\multirow{2}{*}[-0.8mm]{Metric} & \multirow{2}{*}[-0.8mm]{Dataset} & \multirow{2}{*}[-0.8mm]{Methods}& \multicolumn{3}{c|}{\textbf{SAM2.1-B+}}    & \multicolumn{3}{c|}{\textbf{SAM2.1-S}} & \multicolumn{3}{c}{\textbf{SAM2.1-T}}        \\
\cmidrule(){4-12} 
&&& FP & W2A4 & W2A2 & FP & W2A4 & W2A2 & FP & W2A4 & W2A2 \\ 
\midrule

\multirow{14}{*}[0.4mm]{$J\&F$} 
&
\multirow{4}{*}{SA-V Val \cite{ravi2024sam2}} 
&  MinMax~\cite{jacob2018quantization}             & \multirow{4}{*}{78.1}   &   52.3    &   18.7    & \multirow{4}{*}{77.7} & 56.8 &   27.8     & \multirow{4}{*}{75.0}   & 53.5  &  17.9    \\ 
&&  PACT~\cite{choi2018pact}                  &                       &     54.7   &  36.4     &                       &    58.4       &  40.0     &                       &     56.3      &  43.4     \\
&&  T\_LSQ+~\cite{bhalgat2020lsqimprovinglowbitquantization}  &                       &     55.0   &  50.7     &                       &    58.1       &  51.7     &                       &     57.5      &  51.4     \\
&&  \cellcolor[HTML]{F3F3F3}\textbf{Q-SAM}                  &                       &    \cellcolor[HTML]{F3F3F3}\textbf{64.7}   & \cellcolor[HTML]{F3F3F3}\textbf{55.2}     &                       &    \cellcolor[HTML]{F3F3F3}\textbf{65.1}       &  \cellcolor[HTML]{F3F3F3}\textbf{56.0}     &                       &     \cellcolor[HTML]{F3F3F3}\textbf{64.1}      &  \cellcolor[HTML]{F3F3F3}\textbf{54.8}     \\  

\cmidrule(){2-12} 

&
\multirow{4}{*}{SA-V Test \cite{ravi2024sam2}} 
&  MinMax~\cite{jacob2018quantization}             & \multirow{4}{*}{78.2}   &   56.8    &   21.2    & \multirow{4}{*}{76.6} & 59.9 &   29.7     & \multirow{4}{*}{75.0}   & 57.1  &   19.5    \\ 
&&  PACT~\cite{choi2018pact}                  &                       &     59.0   &  37.4     &                       &    60.6       &  43.1    &                       &     59.5      &  45.5     \\
&&  T\_LSQ+~\cite{bhalgat2020lsqimprovinglowbitquantization}  &                       &     59.0   &  53.3     &                       &    60.8       &  54.3     &                       &     60.0      &  55.5     \\
&&  \cellcolor[HTML]{F3F3F3}\textbf{Q-SAM2}                  &                       &    \cellcolor[HTML]{F3F3F3}\textbf{64.3}   & \cellcolor[HTML]{F3F3F3}\textbf{58.3}     &                       &    \cellcolor[HTML]{F3F3F3}\textbf{67.9}       &  \cellcolor[HTML]{F3F3F3}\textbf{59.1}     &                       &     \cellcolor[HTML]{F3F3F3}\textbf{67.1}      &  \cellcolor[HTML]{F3F3F3}\textbf{59.3}     \\

\cmidrule(){2-12} 

&
\multirow{4}{*}{MOSE Val \cite{MOSE}} 
&  MinMax~\cite{jacob2018quantization}             & \multirow{4}{*}{73.7}   &   55.3    &   26.5    & \multirow{4}{*}{73.5} & 59.1 &   33.5     & \multirow{4}{*}{70.9}   & 58.8  &   26.2    \\ 
&&  PACT~\cite{choi2018pact}                  &                       &     58.6   &  40.8     &                       &    60.6       &  43.7     &                       &     60.9      &  44.3     \\
&&  T\_LSQ+~\cite{bhalgat2020lsqimprovinglowbitquantization}  &                       &     56.0   &  52.9     &                       &    61.5       &  53.0     &                       &     60.2      &  54.7     \\
&&  \cellcolor[HTML]{F3F3F3}\textbf{Q-SAM2}                  &                       &    \cellcolor[HTML]{F3F3F3}\textbf{64.1}   & \cellcolor[HTML]{F3F3F3}\textbf{57.1}     &                       &    \cellcolor[HTML]{F3F3F3}\textbf{67.4}       &  \cellcolor[HTML]{F3F3F3}\textbf{57.6}     &                       &     \cellcolor[HTML]{F3F3F3}\textbf{66.8}      &  \cellcolor[HTML]{F3F3F3}\textbf{58.2}     \\

\midrule

\multirow{4}{*}{mAP} 
&
\multirow{4}{*}{COCO2017 \cite{cocodataset}} 
&  MinMax~\cite{jacob2018quantization}             & \multirow{4}{*}{49.6}   &   36.2    &   15.4    & \multirow{4}{*}{48.4} & 37.5 &   20.3     & \multirow{4}{*}{48.0}   & 36.5  &   16.8    \\ 
&&  PACT~\cite{choi2018pact}                  &                       &     38.3   &  25.2     &                       &    38.8       &  27.2     &                       &     38.7      &  28.2     \\
&&  T\_LSQ+~\cite{bhalgat2020lsqimprovinglowbitquantization}  &                       &     39.3   &  34.6     &                       &    40.2       &  31.4     &                       &    38.8      &  32.3     \\
&&  \cellcolor[HTML]{F3F3F3}\textbf{Q-SAM2}                  &                       &    \cellcolor[HTML]{F3F3F3}\textbf{45.8}   & \cellcolor[HTML]{F3F3F3}\textbf{39.2}     &                       &    \cellcolor[HTML]{F3F3F3}\textbf{44.3}       &  \cellcolor[HTML]{F3F3F3}\textbf{36.1}     &                       &     \cellcolor[HTML]{F3F3F3}\textbf{43.8}      &  \cellcolor[HTML]{F3F3F3}\textbf{35.4}     \\

\midrule

\multirow{4}{*}{mIoU} 
&
\multirow{4}{*}{COCO2017 \cite{cocodataset}} 
&  MinMax~\cite{jacob2018quantization}             & \multirow{4}{*}{59.1}   &   39.0    &   25.0    & \multirow{4}{*}{59.7} & 41.2 &   26.9     & \multirow{4}{*}{57.1}   & 40.3  &   26.5    \\ 
&&  PACT~\cite{choi2018pact}                  &                       &     41.2   &  32.0     &                       &    41.9       &  32.6     &                       &     38.7      &  28.2     \\
&&  T\_LSQ+~\cite{bhalgat2020lsqimprovinglowbitquantization}  &                       &     42.1   &  37.9     &                       &    42.6       &  36.4     &                       &    42.5      &  37.3     \\
&&  \cellcolor[HTML]{F3F3F3}\textbf{Q-SAM2}                  &                       &    \cellcolor[HTML]{F3F3F3}\textbf{49.4}   & \cellcolor[HTML]{F3F3F3}\textbf{42.1}     &                       &    \cellcolor[HTML]{F3F3F3}\textbf{47.3}       &  \cellcolor[HTML]{F3F3F3}\textbf{39.6}     &                       &     \cellcolor[HTML]{F3F3F3}\textbf{47.5}      &  \cellcolor[HTML]{F3F3F3}\textbf{39.8}     \\

\bottomrule
\end{tabular}
}

\caption{Quantitative results on Video Object Segmentation (VOS) and Instance Segmentation. Our method is highlighted in grey. \textbf{Bold} indicates the best result for each configuration.}
\label{tab:main-results}
\vspace{-1mm}
\end{table*}

\subsection{Experimental Setup}
\subsubsection{Task, Dataset, and Metrics}
Our experimental setup resembles the one in the original SAM2 work \cite{ravi2024sam2}. We apply the VRC and train the models on a subset of the SA-1B \cite{kirillov2023segany} and SA-V \cite{ravi2024sam2} datasets. The SA-1B dataset is composed of 11M images and more than 1B masks, while SA-V contains 51k videos and 643K spatio-temporal segmentation masks. To reduce training time, we use a smaller subset of the dataset and still obtained high accuracy, demonstrating that we achieve performance without full retraining. From the original data, we subsample at random \(\sim 8\%\) and \(\sim 45\%\), respectively. 

We evaluate our models on the semi-supervised Visual Object Segmentation (VOS) task, which involves segmenting a target object across a video given its ground-truth mask in the first frame. Following the SAM2 evaluation setup, we use SA-V val, SA-V test (both subsets of SA-V), and MOSE val from the MOSE dataset~\cite{MOSE}, reporting $J\&F$ accuracy~\cite{Pont-TusetPCASG17}. For instance segmentation, we evaluate on the COCO2017 validation set~\cite{cocodataset}, using mean Intersection over Union (mIoU) across all 5,000 images. Moreover, following the same evaluation as PTQ4SAM~\cite {PTQ4SAMLv}, we use the predicted boxes generated from DINO~\cite{zhang2022dinodetrimproveddenoising} as input prompt for our model, in order to calculate the mean Average Precision (mAP) on COCO2017.

\subsubsection{Implementation}
Our starting models are the checkpoints available from the SAM2 repository \cite{sam2github2024}, named \texttt{sam2.1\_hiera\_tiny} (T), \texttt{sam2.1\_hiera\_small} (S), \texttt{sam2.1\_hiera\_base\_plus} (B+). The implementation of the QAT pipelines is based on PyTorch 2.6.0 and CUDA 12.4, and all experiments are performed on NVIDIA A100 80GB GPUs. 
The VRC is then performed using 50 input images, chosen uniformly at random from the SA-1B dataset. We forward the images through the original image encoder and record the activations. After stacking the tensors, we obtain the calibrated weight matrices by applying equation~\ref{eq:regularization} with $\lambda_0=2.0$.
To assess our method, we compare Q-SAM2 against the classical MinMax method~\cite{jacob2018quantization}, PACT~\cite{choi2018pact}, and LSQ+~\cite{bhalgat2020lsqimprovinglowbitquantization}. Since standard LSQ+ (and LSQ~\cite{esser2019lsq}) fails to converge with per-channel quantization~\cite{li2022mqbenchreproducibledeployablemodel}, we implement a Tuned LSQ+ (T\_LSQ+) baseline. This involves restricting the original LSQ+ to only activations quantization, tuning the learning rate for the step size $s$ and offset $z$, increasing observer calibration samplings, and employing per-channel MinMax for weights. To ensure a fair comparison, we evaluate LSQ+ against Q-SAM2 in the W4A4 per-tensor schema. Even in this setting, Q-SAM2 maintains a significant lead, outperforming LSQ+ by 9.7 ppt (VOS J\&F) and 7.7 ppt (COCO mIoU). Full details, training pipelines, parameters, and losses are provided in Appendix~\ref{sec:train_results}.
Finally, we validate the stability of our training pipeline, observing a marginal standard deviation of 0.26 ppt in mIoU across independent runs.

\subsection{Ultra-Low-Bit Quantization Performance}
\label{sec:results_ul}
Table \ref{tab:main-results} summarizes the results of our experiments. The results show that our method consistently outperforms the other QAT algorithms by a large margin across each task, dataset, encoder size, and precision configuration. For the B+ encoder, Q-SAM2 demonstrates substantial improvement over the best baseline, achieving average gains on the video benchmark of 7.0 ppt (W2A4) and 4.6 ppt (W2A2). For instance segmentation, gains are 6.5 ppt (W2A4) and 4.6 ppt (W2A2) in mAP, and 7.3 ppt (W2A4) and 4.2 ppt (W2A2) in mIoU.
The performance margin is reduced for the small and tiny encoders. This narrowing is expected because smaller models are inherently easier to quantize due to fewer outliers and narrower weight and activation distributions compared to the B+ configuration. This observation is attributed to the fact that the simpler distributions of the S and T encoders significantly boost the absolute performance of existing QAT baselines, thereby narrowing the gap with Q-SAM2. While T\_LSQ+ substantially reduces the performance difference in the aggressive W2A2 configuration, Q-SAM2 retains a significant lead. While the tiny configuration yields our smallest performance margin relative to baselines, representing the worst-case scenario for our gains, we still achieve substantial improvements averaging 3.6 ppt on the video benchmark, 3.1 ppt in mAP, and 2.5 ppt in mIoU against T\_LSQ+.  

In a second experiment we evaluate our Q-SAM2 against post-training quantization methods, specifically PTQ4SAM~\cite{PTQ4SAMLv}, BRECQ~\cite{li2021brecq}, and QDROP~\cite{wei2023qdroprandomlydroppingquantization}, on the instance segmentation task. As visualized in Figure~\ref{fig:ptq_comparison}, our lightweight Q-SAM2 W2A4\_B+ model achieves a 1.9 ppt mAP gain over the more than $\times8$ larger (in MB) PTQ4SAM-H W4A4 model. The complete comparison is reported in Appendix~\ref{sec:train_results}.

\subsection{Ablation Studies}
 \begin{table}[htbp]

    \centering
    \tiny
    \caption{Ablation study on VRC and LSC for B+ model. We report mAP for instance segmentation on COCO2017 dataset. }

    \label{tab:ablation_main}
    
    \resizebox{0.48\textwidth}{!}{
    \begin{tabular}{ccc|c}

        \toprule

        \textbf{Dataset} & \textbf{VRC $(\lambda_0=2.0)$} & \textbf{LSC} & \textbf{W2A2} \\

        \midrule

        \multirow{4}{*}{COCO2017~\cite{cocodataset}} & \myxmark & \myxmark & 15.4 \\

        & \mycheckmark & \myxmark & 18.3 \\

        & \myxmark & \mycheckmark & 37.1 \\

        &   \cellcolor[HTML]{F3F3F3}\mycheckmark & \cellcolor[HTML]{F3F3F3}\mycheckmark & \cellcolor[HTML]{F3F3F3}\textbf{39.2} \\


        \bottomrule

    \end{tabular}
}

\end{table}

\begin{table}[htbp]

    \centering
    \tiny

    \caption{Ablation studies on the VRC for weight-only quantization for B+ encoder. The mAP for the image segmentation on COCO2017 and J\&F for the VOS on SA-V val are reported.}

    \label{tab:abl_vrc}
    \resizebox{0.48\textwidth}{!}{
    \begin{tabular}{ccc|c|c}

        \toprule

        \textbf{Dataset} & \textbf{Method} & \textbf{VRC $(\lambda_0)$} & \textbf{W3} & \textbf{W4} \\
        \midrule
        \multirow{6}{*}{COCO2017~\cite{cocodataset}} & \multirow{3}{*}{HQQ~\cite{badri2023hqq}}
        &  \myxmark & 24.9 & 47.5 \\
        & &\cellcolor[HTML]{F3F3F3}\mycheckmark {\color{black}$(1.5)$} & \cellcolor[HTML]{F3F3F3}29.4 & \cellcolor[HTML]{F3F3F3}\textbf{47.6} \\
        & &\cellcolor[HTML]{F3F3F3}\mycheckmark {\color{black}$(2.0)$} & \cellcolor[HTML]{F3F3F3}\textbf{29.6} & \cellcolor[HTML]{F3F3F3}46.0 \\
        \cmidrule(){2-5} 
        & \multirow{3}{*}{MinMax~\cite{jacob2018quantization}}
        &  \myxmark & 15.7 & 46.3\\
        & & \cellcolor[HTML]{F3F3F3}\mycheckmark {\color{black}$(1.5)$} & \cellcolor[HTML]{F3F3F3}19.6 & \cellcolor[HTML]{F3F3F3}\textbf{46.9} \\
        & & \cellcolor[HTML]{F3F3F3}\mycheckmark {\color{black}$(2.0)$} & \cellcolor[HTML]{F3F3F3}\textbf{20.3} & \cellcolor[HTML]{F3F3F3}45.2 \\

        \midrule
        \multirow{6}{*}{SA-V val \cite{ravi2024sam2}} & \multirow{3}{*}{HQQ~\cite{badri2023hqq}}
        &  \myxmark & 38.1 & \textbf{71.0} \\
        & &\cellcolor[HTML]{F3F3F3}\mycheckmark {\color{black}$(1.5)$} & \cellcolor[HTML]{F3F3F3}41.2 & \cellcolor[HTML]{F3F3F3}\textbf{71.0} \\
        & &\cellcolor[HTML]{F3F3F3}\mycheckmark {\color{black}$(2.0)$} & \cellcolor[HTML]{F3F3F3}\textbf{43.0} & \cellcolor[HTML]{F3F3F3}67.6 \\
        \cmidrule(){2-5}
        & \multirow{3}{*}{MinMax~\cite{jacob2018quantization}}
        &  \myxmark & 34.4 & 69.2 \\
        & & \cellcolor[HTML]{F3F3F3}\mycheckmark {\color{black}$(1.5)$} & \cellcolor[HTML]{F3F3F3}36.6 & \cellcolor[HTML]{F3F3F3}\textbf{70.9} \\
        & & \cellcolor[HTML]{F3F3F3}\mycheckmark {\color{black}$(2.0)$} & \cellcolor[HTML]{F3F3F3}\textbf{39.5} & \cellcolor[HTML]{F3F3F3}64.4 \\

        \bottomrule
    \end{tabular}
    }
\end{table} 
\begin{figure*}[htb!]
  \centering
  
    \begin{adjustbox}{width=\textwidth}
    \begin{tabular}{>{\centering\arraybackslash}m{0.25\textwidth} 
                    >{\centering\arraybackslash}m{0.25\textwidth} 
                    >{\centering\arraybackslash}m{0.25\textwidth}
                    >{\centering\arraybackslash}m{0.25\textwidth}}
        \small Full Precision & \small T-LSQ+~\cite{bhalgat2020lsqimprovinglowbitquantization} & \small PACT~\cite{choi2018pact} & \small \textbf{Q-SAM2 (ours)}
    \end{tabular}
    \end{adjustbox}

  \vspace{0.5em} 
    \includegraphics[width=1.\textwidth]{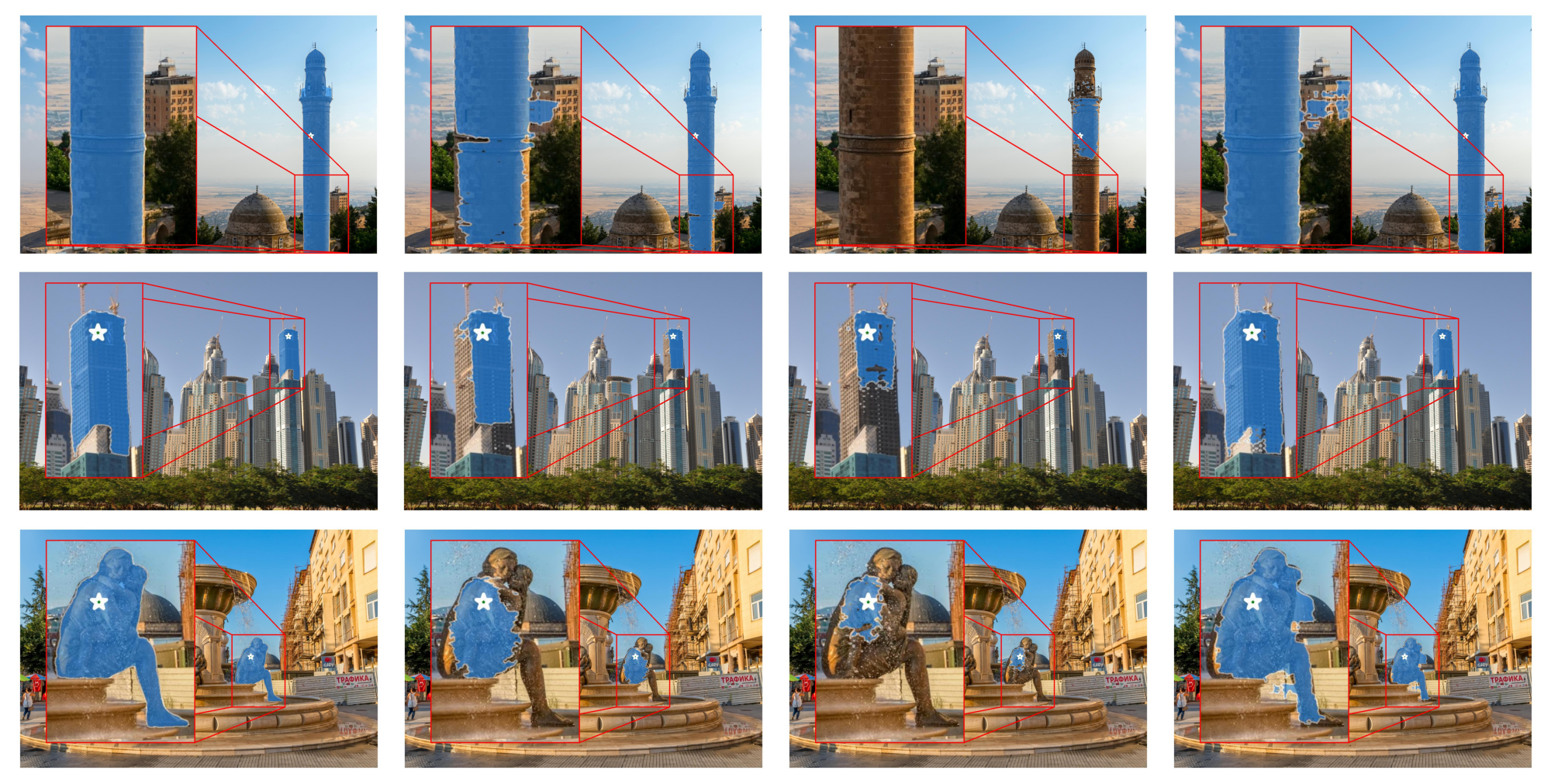} 
  \caption{Qualitative results on W2A2 configuration for B+ encoder on promptable instance segmentation task.}
  \label{fig:visual_results}
\end{figure*}
Table~\ref{tab:ablation_main} reports the contributions of VRC and LSC in trained networks. While the largest contribution stems from the LSC's management of activation outliers, VRC yields a 2.1 ppt gain. This improvement is statistically significant, as it substantially exceeds the observed training standard deviation of 0.26 ppt.
The second ablation study evaluates the VRC calibration procedure in a post-training quantization setting using the B+ model with 3- and 4-bit weight-only quantization. We test VRC with two values of $\lambda_0$ on two methods: standard MinMax~\cite{jacob2018quantization} and the Hessian-aware HQQ~\cite{badri2023hqq}. As detailed in Table~\ref{tab:abl_vrc}, VRC consistently improves performance across both strategies. The gain is most significant in the 3-bit configuration, achieving a 4.6 ppt mAP gain for MinMax. In the 4-bit setting, the improvement is marginal ($\lambda_0=1.5$ offers a slight gain), confirming the expected trend: the contribution of VRC decreases as the bit-width becomes sufficient to represent weights accurately.

\subsection{Qualitative Results}
Figure~\ref{fig:visual_results} presents a qualitative comparison of Q-SAM2 versus full-precision SAM2.1~\cite{ravi2024sam2} and quantized baselines PACT~\cite{choi2018pact} and T-LSQ+~\cite{bhalgat2020lsqimprovinglowbitquantization}. Visually confirming our quantitative advantage, Q-SAM2 demonstrates markedly improved segmentation quality. PACT consistently exhibits high levels of artifacts across all images and substantially fails to segment the input prompt. In contrast, T\_LSQ+ fares marginally better, achieving only partial segmentation in some instances but invariably introducing noticeable artifacts. Our Q-SAM2 yields segmentations closest to the full-precision model. Additional visual results and a video example are available in Appendix~\ref{sec:qres}.

\section{Conclusions}
\label{sec:conclusions}
We presented Q-SAM2, to the best of our knowledge the first quantized Segment Anything Model 2. Our method combined VRC, which reduces weight variance, and LSC, a robust QAT procedure that learns to clip outliers. Q-SAM2 significantly outperformed strong baselines on VOS and instance segmentation. Across all encoder sizes and quantization schemes evaluated, we observed average gains of 5.5~ppt J\&F on the video benchmark, 4.7~ppt in mAP, and 4.5~ppt in mIoU, while qualitative results confirmed its effectiveness in preserving segmentation quality. 

Future work could investigate the generalization of VRC and LSC to other architectures. A particularly promising direction involves exploring an adaptive VRC with a layer-specific $\lambda$ parameter, which enables a more fine-grained calibration to further optimize the accuracy-compression trade-off.

\section*{Acknowledgement}
We acknowledge funding from the European Union’s Horizon 2020 research and innovation programme under the Marie Skłodowska-Curie actions HORIZON-MSCA-2022-DN-01 call (Grant agreement ID: 101119554).
{
    \small
    \bibliographystyle{ieeenat_fullname}
    \bibliography{main}
}

\clearpage
\section{Technical Appendices and Supplementary Material}
This appendix provides more details to support the findings presented in the main paper.
In Section~\ref{sec:calibration}, we analyze the bias-variance trade-off introduced by the regularization hyperparameter \(\lambda\) in the VRC, Section~\ref{sec:train_results} outlines our training configuration and other results. In particular  Section~\ref{sec:training_loss_plots} presents training loss curves, and Section~\ref{sec:training_sam2_comparison} discusses the differences between our training procedure and the original SAM2 model~\cite{ravi2024sam2}. Section~\ref{sec:training_lsq} the LSQ+~\cite{bhalgat2020lsqimprovinglowbitquantization} QAT algorithm is compared with our Q-SAM2 in a per-tensor schema, and in Section~\ref{sec:training_ptq} we compare our performance with PTQ algorithm applied on SAM~\cite{kirillov2023segany}. We present insignts on the weight and activation distributions of SAM2 in Section~\ref{sec:distributions}. Finally, Section~\ref{sec:qres} includes qualitative segmentation results to illustrate the effectiveness of our approach in both static images and a video.

\subsection{Variance-Reduced Calibration Details}
\label{sec:calibration}
\begin{figure*}[hbt]
  \centering
  \begin{subfigure}[t]{0.45\textwidth}
    \centering
      \includegraphics[width=\textwidth]{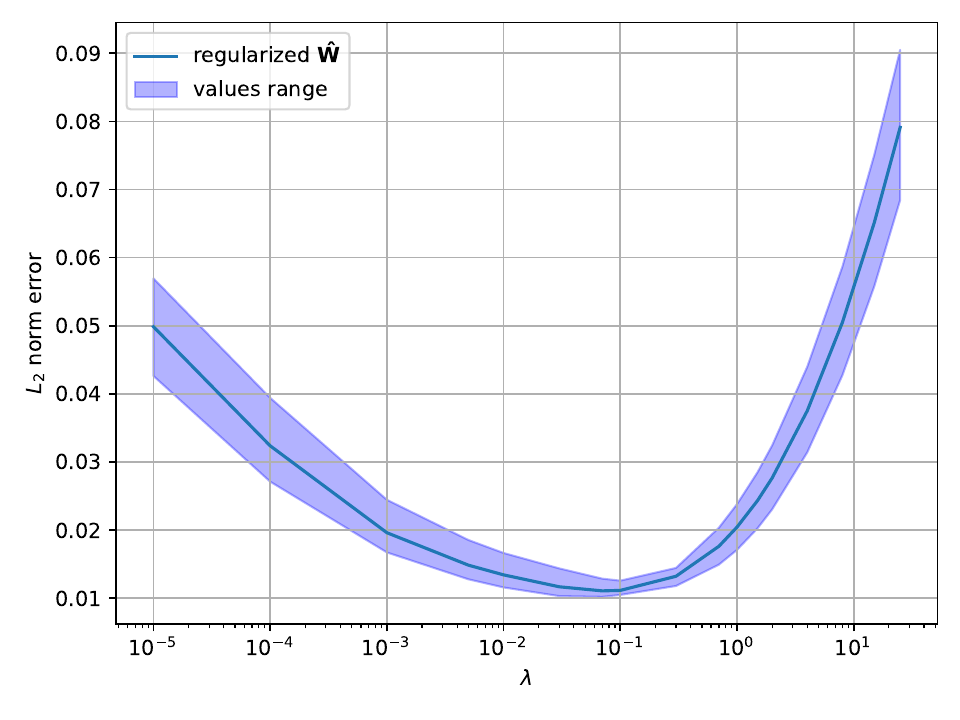}
      \caption{\(L_2\) error over different \(\lambda\)}
      \label{fig:l2_error_calib}
  \end{subfigure}
  \hfill
  \begin{subfigure}[t]{0.45\textwidth}
    \centering
  \includegraphics[width=\textwidth]{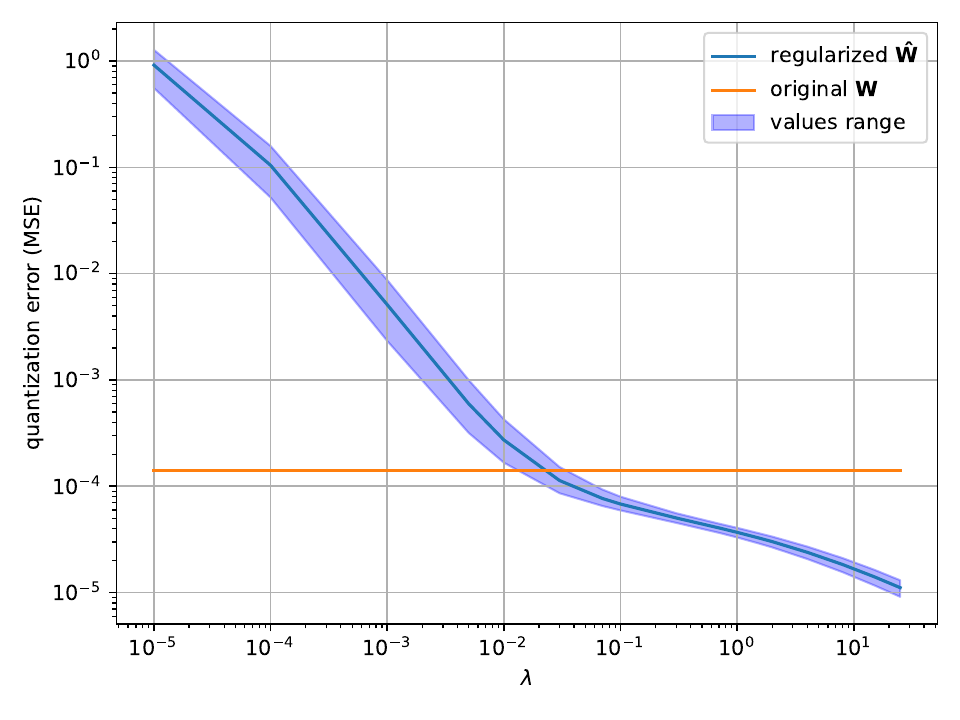}
  \caption{Quantization error (MSE) over different \(\lambda\)}
  \label{fig:quant_error_calib}
  \end{subfigure}

  \caption{Calibration error for the linear layer \textit{cut.blocks.21.mlp.layers.1} from the image encoder of the B+ model. In subfigure (a), we compare the output of the original linear layer with the output produced using the calibrated weights \(\mathbf{\hat{W}}_\lambda\), measured via the \(L_2\) norm, across different values of the hyperparameter \(\lambda\). In subfigure (b), we report the quantization error of \(\mathbf{\hat{W}}_\lambda\) for varying \(\lambda\), relative to the quantization error of the original weights \(\mathbf{W}\).}
  \label{fig:calib_supp}
\end{figure*}
To better analyze the behavior of our calibration method introduced in Section~\ref{sec:VRC}, we implement Equation~\ref{eq:regularization} within the linear layer \textit{cut.blocks.21.mlp.layers.1} of the image encoder in the B+ model. Calibration is performed by applying the regularized pseudoinverse formulation to batches of 50 images, chosen uniformly at random, while systematically varying the regularization hyperparameter \(\lambda\). For each value of \(\lambda\), we compute a calibrated weight matrix \(\mathbf{\hat{W}}_\lambda\in \mathbb{R}^{d_{\text{out}}\times d_{\text{in}}}\).

To evaluate the effect of calibration, we compare the outputs of the original and calibrated layers using a separate evaluation batch of 10 images, denoted as \(\mathbf{X}_{\text{eval}}\). The output discrepancy is quantified using the \(L_2\) norm between the original projection \(\mathbf{X}_{\text{eval}}\mathbf{W}^\top\) and the calibrated output \(\mathbf{X}_{\text{eval}}\mathbf{\hat{W}}^\top_\lambda\). This metric provides a direct measure of the reconstruction error induced by the calibration process.

Figure~\ref{fig:l2_error_calib} illustrates how the \(L_2\) error varies with \(\lambda\), revealing a clear bias-variance trade-off. For small regularization values (\(\lambda < 10^{-3}\)), the error is high and exhibits large variance across different calibration batches, indicating sensitivity to batch selection due to ill-conditioned activations. This suggests that when regularization is too weak, the inversion step amplifies noise, degrading output fidelity. On the opposite end, overly strong regularization (\(\lambda > 1\)) significantly alters \(\mathbf{\hat{W}}_\lambda\), reducing its ability to generalize and match the behavior of the original layer. In the intermediate range (\(10^{-3} \leq \lambda \leq 1\)), the reconstruction error is minimized, indicating an optimal trade-off that stabilizes calibration without overfitting to the batch.

To further assess the impact of calibration on quantization, we analyze the quantization error of \(\mathbf{\hat{W}}_\lambda\), as shown in Figure~\ref{fig:quant_error_calib}. The error is computed as \[\frac{1}{d_{\text{out}}d_{\text{in}}}\|\mathbf{\hat{W}}_\lambda - \mathbf{\hat{W}}_{\lambda_Q}\|^2_F ,\] where \(\mathbf{\hat{W}}_{\lambda_Q}\) is the quantized version of the calibrated weight matrix using uniform quantization. The orange line in the figure represents the quantization error when using the original weight matrix \(\mathbf{W}\) instead of \(\mathbf{\hat{W}}_\lambda\). We observe that counterintuitively for low \(\lambda\) values, the quantization error is extremely high due to the poor conditioning of \(\mathbf{\hat{W}}_\lambda\), which leads to numerical instability and large weight magnitudes. As \(\lambda\) increases, particularly beyond \(5 \times 10^{-2}\), the quantization error of the calibrated weights becomes consistently lower than that of the original weights. This suggests that regularization not only improves numerical stability, but also produces weight matrices that are better structured for quantization, exhibiting smaller dynamic ranges.

The selection of \(\lambda\) is not straightforward. One option is to choose the value that minimizes the reconstruction error in Figure~\ref{fig:l2_error_calib}, which provides a good match to the original layer outputs. However, this choice does not necessarily lead to the lowest quantization error, and its impact on initialization quality for quantized training may be limited. Conversely, setting \(\lambda \approx 1\) improves the quantization behavior but can result in higher \(L_2\) output error, compromising the layer's ability to generalize. This highlights the need to balance fidelity to the original outputs with improved robustness under quantization when selecting the regularization strength.
 
To select an appropriate value of \(\lambda\) in practice, we adopt a rule based on the singular value spectrum of the input matrix. If the activation matrix is ill-conditioned (i.e., with a condition number greater than \(10^2\)) but not effectively rank deficient, we set \(\lambda\) to five times the smallest non-zero singular value. In the case of low-precision formats such as \texttt{bfloat16}, matrices with condition numbers exceeding \(10^3\) can already exhibit numerical instability. When the matrix is more severely ill-conditioned, with a condition number significantly greater than \(10^3\), it is likely to be effectively rank deficient. In this case, we identify the smallest singular value \(\sigma_{*}\) such that \(\sigma_{\max} / \sigma_{*} \leq 10^3\), and set \(\lambda= \lambda_0\sigma_{*}\).

As a general rule, depending on the conditioning characteristics of the specific network and input data, \(\lambda_0\) is chosen between two and five times the selected singular value. This strategy stabilizes the pseudoinverse computation while preserving the most meaningful directions in the activation space.

\begin{figure*}[htb]
  \centering
  \begin{subfigure}[t]{0.45\textwidth}
    \centering
      \includegraphics[width=\textwidth]{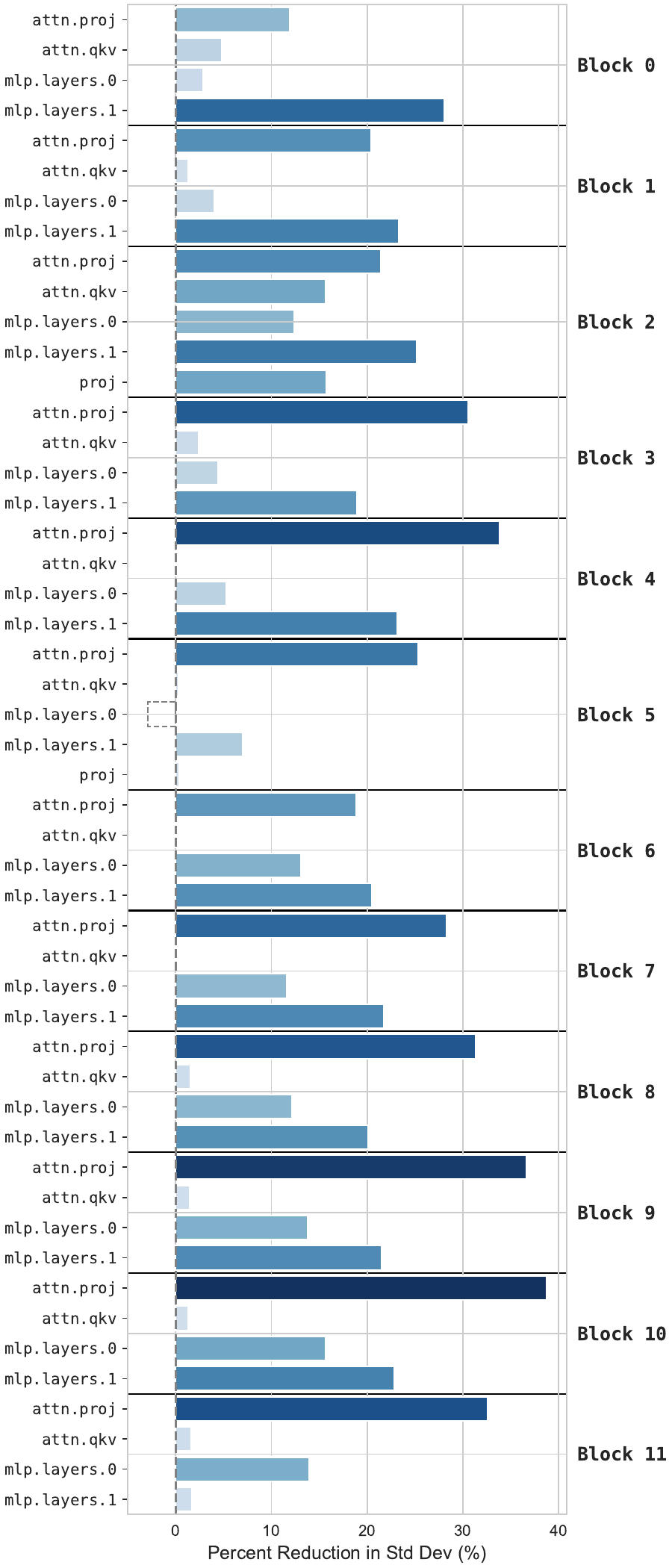}
  \end{subfigure}
  \hfill
  \begin{subfigure}[t]{0.45\textwidth}
    \centering
  \includegraphics[width=\textwidth]{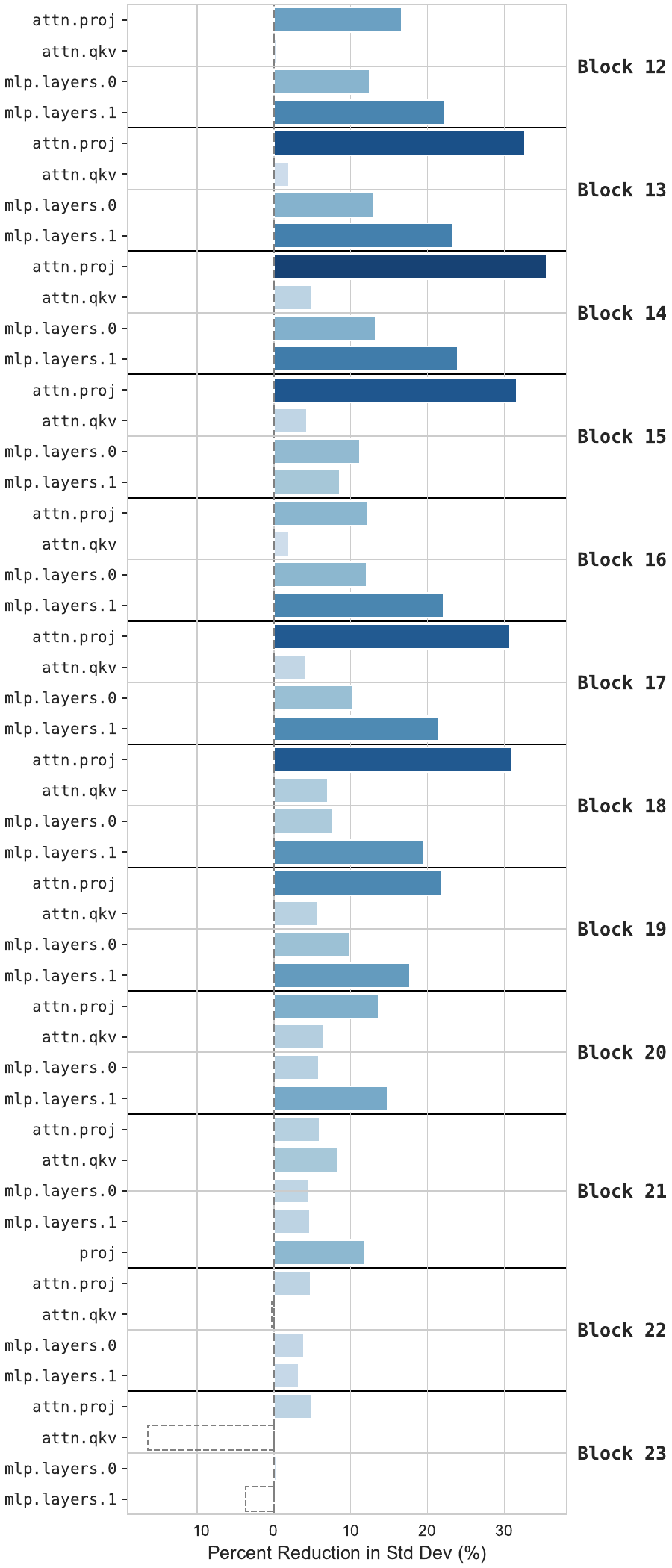}
  \end{subfigure}

  \caption{Impact of VRC on SAM2.1-B+ image encoder weight distributions for $\lambda_0=2.0$. VRC achieves a reduction up to 38.7\%, with an average value of 13.2\%.}
  \label{fig:st_reduction}
\end{figure*}

In Figure~\ref{fig:st_reduction} we present the percentage reduction in standard deviation of all the layers after applying VRC to the B+ encoder when $\lambda_0=2.0$. In several blocks, some layers achieve a $\sigma$ reduction exceeding 30\%, which significantly lowers the initial quantization error. As detailed in Section~\ref{sec:VRC}, the chosen $\lambda_0$ value may be insufficient for certain layers; we observe that three layers specifically fallback to their original uncalibrated state.

\subsection{Training and Other Results}
\label{sec:train_results}
\subsubsection{Training Parameters}
\label{sec:training}
\begin{table*}[htb]
\centering
\begin{tabularx}{\textwidth}{YY}
\toprule
Configuration & Value \\
\midrule
data &  \textbf{SA-1B(8\%)  SA-V(45\%)}\\
resolution & 1024\\
precision & bfloat16\\
epochs & \textbf{1} \\
optimizer & AdamW\\
optimizer momentum & \(\beta_1, \beta_2=0.9, 0.999\)\\
gradient & clipping type: \(L_2\), max: 0.1\\
weight decay & 0.1\\
learning rate (lr) & \textbf{img. enc.: }\(\mathbf{1e-5}\), other: \(3.0e-4\)\\
lr schedule & cosine\\
warmup & \textbf{no warmup}\\
layer-wise decay & 0.8 (T, S), 0.9 (B+)\\
image augmentation & hflip, resize to 1024 (square)\\
video augmentation&  hflip, affine (deg: 25, shear: 20), colorjitter (b: 0.1, c: 0.03, s: 0.03,h: null), grayscale (0.05), per frame colorjitter (b: 0.1, c: 0.05, s:0.05, h: null), mosaic-2×2 (0.1)\\
batch size & 256\\
drop path & 0.1 (T, S), 0.2 (B+)\\
mask losses (weight)&  focal (20), dice (1)\\
IoU loss (weight) & $L_1$ (1)\\
occlusion loss (weight) & cross-entropy (1)\\
max. masks per frame & \textbf{image: 60}, video: 3\\
\# correction points & 7\\
global attn. blocks &  5-7-9 (T), 7-10-13 (S), 12-16-20 (B+)\\
\bottomrule
\end{tabularx}%
\caption{Hyperparameters and details of our Q-SAM2 and baselines QAT training for the three image encoder sizes B+,S,T.}
\label{tab:hyperparams}
\end{table*}
To ensure fairness, we used a consistent training and evaluation setup for both our Q-SAM2 and the baseline approaches Minmax~\cite{jacob2018quantization}, PACT~\cite{choi2018pact}, and T\_LSQ+~\cite{bhalgat2020lsqimprovinglowbitquantization}. The training procedure follows the one proposed in the SAM2 \cite{ravi2024sam2} paper, where the authors referred to the "full training" step. The details are reported in Table~\ref{tab:hyperparams} where we highlight the configurations and hyperparameters that are different from the original work. All trained models use the same image and video data, with a fixed random seed to ensure reproducibility. 

Compared to the original SAM2 training setup, we introduce several modifications to accommodate hardware constraints and reduce training time. First, we reduce the maximum number of masks per frame from 64 to 60 to fit within GPU memory limits on an NVIDIA A100 80GB. We also skip the warm-up phase, reduce the amount of input data, and limit training to a single epoch. 

All training experiments are conducted using 8 NVIDIA A100 80GB GPUs, with a total training time of less than 35 hours for the \textit{B+} architecture.

We adjust the main learning rate from $4 \times 10^{-5}$ to $1 \times 10^{-5}$ and decrease the effect of the gradient scaler by halving its initial scale value. For the T\_LSQ+ training, we reduce the scale and zero-point learning rate to $5 \times 10^{-8}$ to stabilize parameter convergence. 

The initial value for the LSC parameter $k$ (Section~\ref{sec:LSC}) is set to $k=2.5$ for B+ weights and activations, as well as for S and T activations. We use $k=2.8$ for the S and T weights. Finally, for observer calibration before QAT, we infer 300 images for Q-SAM2, PACT, and MinMax, but increase this to 900 images for T\_LSQ+ to capture its activation distributions more effectively.

\subsubsection{Losses Plots}
\label{sec:training_loss_plots}

\label{sec:loss_plot}
\begin{figure*}
  \centering
  \includegraphics[width=\textwidth]{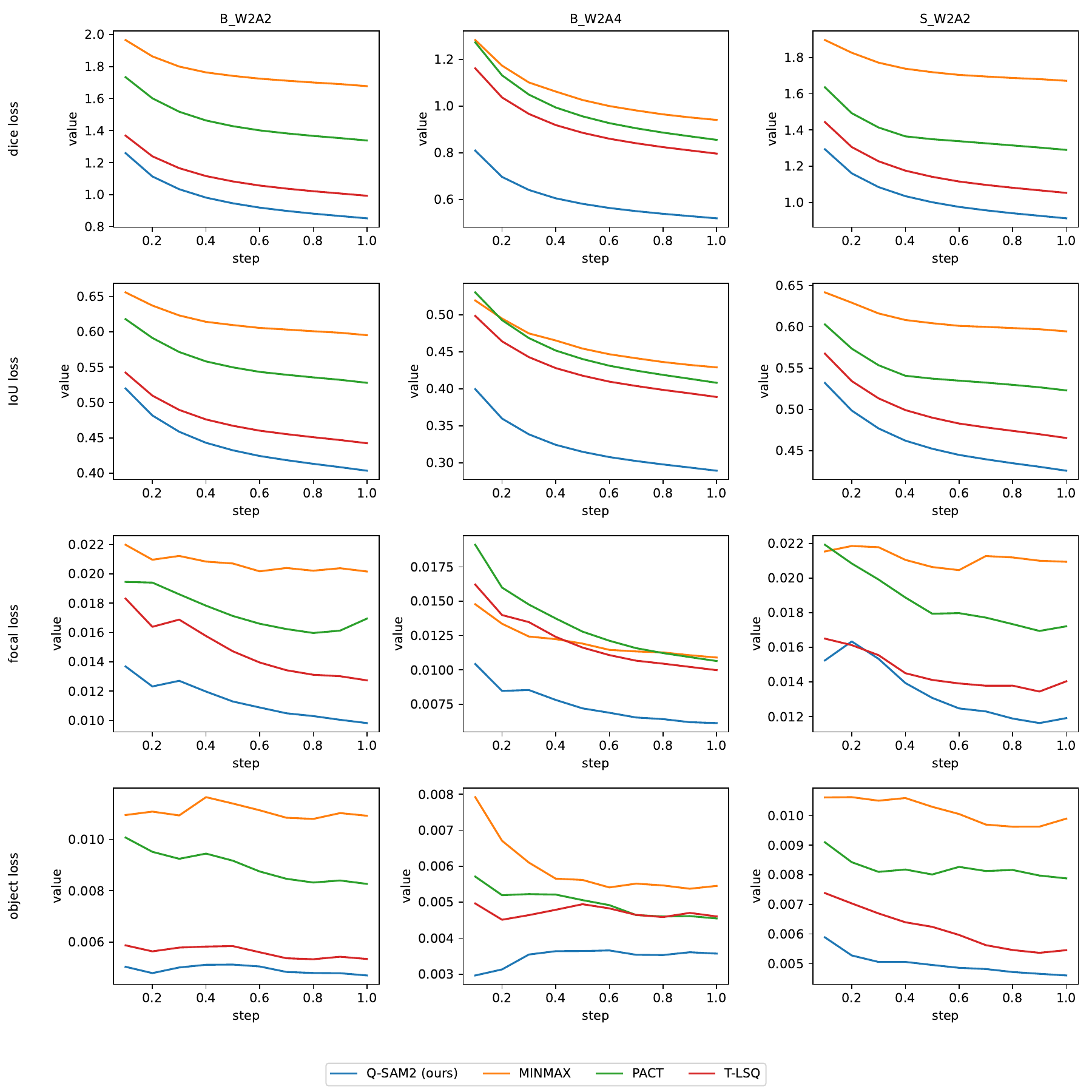}
  \caption{Training loss curves for image encoder size B+, S under two configurations W2A4 and W2A2. Each sub-plot contains a comparison between Q-SAM2 and the baselines for the four losses: dice, IoU, focal, and occlusion. Our solution outperforms the baselines in all losses for all configurations. }
  \label{fig:losses1}
\end{figure*}

\begin{figure*}
  \centering
  \includegraphics[width=\textwidth]{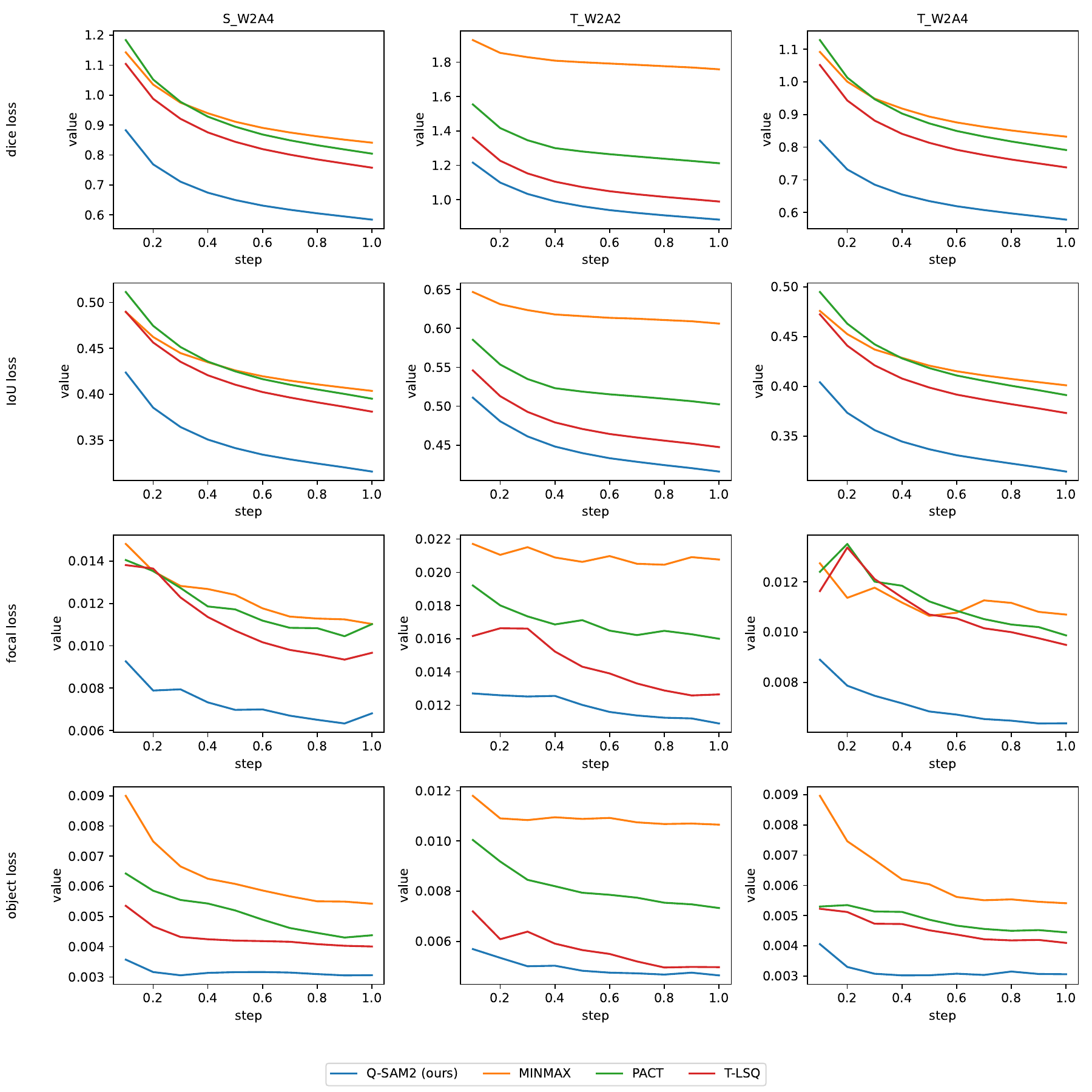}
  \caption{Continuation of the training loss curves from Figure \ref{fig:losses1} for image encoder size S, T under two configurations W2A4 and W2A2. Each sub-plot contains a comparison between Q-SAM2 and the baselines for the four losses: dice, IoU, focal, and occlusion. Our solution outperforms the baselines in all losses all configurations.}
  \label{fig:losses2}
\end{figure*}

We adopt the standard loss of SAM2 \cite{ravi2024sam2} as \(\mathcal{L}_{\text{SAM2}}\), which is composed of a linear combination of focal and dice losses for mask prediction, mean absolute error (MAE) loss for IoU prediction, and cross-entropy loss for object prediction. The weights are set in a fixed ratio of \(20:1:1:1\), respectively. 

Figures~\ref{fig:losses1} and~\ref{fig:losses2} present the loss curves for the W2A4 and W2A2 quantization settings. Specifically, Figure~\ref{fig:losses1} shows the results for the B+ model and one configuration of the S model (W2A2), while Figure~\ref{fig:losses2} includes the second S configuration (W2A4) together with both configurations for the T model.

In all settings and across all loss components, Q-SAM2 demonstrates lower training losses compared to the MinMax, PACT, and T\_LSQ+ baselines. In addition, our model converges faster as the difference in loss values, particularly for the dice and IoU components, increases progressively with each training step.

\subsubsection{Comparison with SAM2 training pipeline}
\label{sec:training_sam2_comparison}
The SAM2 is trained in two main stages followed by a fine-tuning procedure. The first stage, pre-training, involves training on static images using the SA-1B dataset. The second stage, called full training, uses a combination of images from SA-1B, videos from SA-V, and additional internal data that is not publicly available. Finally, to improve segmentation performance on long video sequences, the authors fine-tune the model by increasing the number of input video sequences from 8 to 16. To accommodate this within GPU memory constraints, the image encoder is frozen during this step.
In our work, we reproduce the second stage, called "full training", focusing on a full architecture training with static image and video data while reducing both the data volume and the number of training steps with respect to the original paper. As demonstrated in Section~\ref{sec:training_loss_plots}, our method learns faster than baselines and achieves accurate results with fewer data, thus reducing training time. However, a direct comparison with the original SAM2 model remains difficult due to differences in training procedures. We plan to explore a fully aligned comparison with the complete SAM2 pipeline in future work.

\subsubsection{Comparison with LSQ}
\label{sec:training_lsq}
\begin{table}[htbp]

    \centering
    \tiny

    \caption{Evaluation of the semi-supervised VOS and instance segmentation tasks for the B+ model and W4A4 (per-tensor) configuration. Our soulution is compared with the LSQ+~\cite{bhalgat2020lsqimprovinglowbitquantization} QAT algorithm.}

    \label{tab:comparison_lsq}
    \resizebox{0.48\textwidth}{!}{
    \begin{tabular}{ccc|c|c}

        \toprule

        \textbf{Metric} & \textbf{Dataset} & \textbf{Method} & \textbf{FP} & \textbf{W4TA4} \\
        \midrule
        \multirow{6}{*}[-1.3mm]{J\&F} & \multirow{2}{*}{SA-V Val~\cite{ravi2024sam2}} & LSQ+~\cite{bhalgat2020lsqimprovinglowbitquantization} & \multirow{2}{*}{78.1} & 56.4\\
         && \cellcolor[HTML]{F3F3F3}\textbf{Q-SAM2} & & \cellcolor[HTML]{F3F3F3}\textbf{65.2}\\
         \cmidrule(){2-5} 
         & \multirow{2}{*}{SA-V Test~\cite{ravi2024sam2}} & LSQ+~\cite{bhalgat2020lsqimprovinglowbitquantization} & \multirow{2}{*}{78.2} & 56.6\\
         && \cellcolor[HTML]{F3F3F3}\textbf{Q-SAM2} & & \cellcolor[HTML]{F3F3F3}\textbf{66}\\
         \cmidrule(){2-5} 
         & \multirow{2}{*}{MOSE Val~\cite{MOSE}} & LSQ+~\cite{bhalgat2020lsqimprovinglowbitquantization} & \multirow{2}{*}{73.7} & 56.6\\
         && \cellcolor[HTML]{F3F3F3}\textbf{Q-SAM2} & & \cellcolor[HTML]{F3F3F3}\textbf{67.5}\\

        \midrule
        \multirow{2}{*}{mAP} & \multirow{2}{*}{COCO2017~\cite{cocodataset}} & LSQ+~\cite{bhalgat2020lsqimprovinglowbitquantization} & \multirow{2}{*}{49.6} & 41.1\\
         && \cellcolor[HTML]{F3F3F3}\textbf{Q-SAM2} & & \cellcolor[HTML]{F3F3F3}\textbf{47.9}\\
         \midrule
        \multirow{2}{*}{mIoU} & \multirow{2}{*}{COCO2017~\cite{cocodataset}} & LSQ+~\cite{bhalgat2020lsqimprovinglowbitquantization} & \multirow{2}{*}{59.1} & 44.9\\
         && \cellcolor[HTML]{F3F3F3}\textbf{Q-SAM2} & & \cellcolor[HTML]{F3F3F3}\textbf{52.6}\\
        \bottomrule
    \end{tabular}
    }
\end{table} 
\begin{figure*}[htb!]
  \centering
  \includegraphics[width=\textwidth]{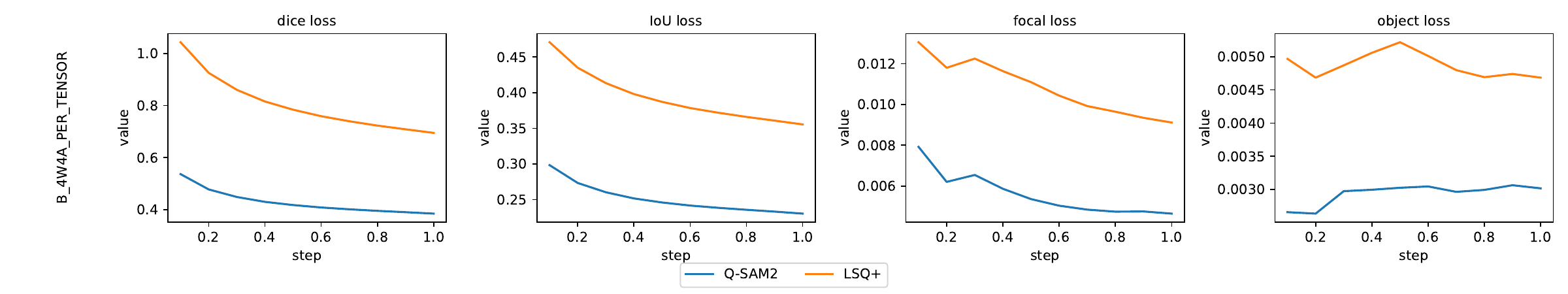}
  \caption{Training loss curves for image encoder size B+ under W4A4 (per-tensor) configurations. Each sub-plot contains a comparison between Q-SAM2 and the LSQ+~\cite{bhalgat2020lsqimprovinglowbitquantization} method for the four losses: dice, IoU, focal, and occlusion.}
  \label{fig:lsq_loss}
\end{figure*}
We compare Q-SAM2 with LSQ+~\cite{bhalgat2020lsqimprovinglowbitquantization}, a widely adopted and robust QAT algorithm known to learn quantization step sizes and zero point via gradient-based optimization. Although LSQ+ has shown strong results on various tasks, Table~\ref{tab:comparison_lsq} shows that Q-SAM2 significantly outperforms on all video and image benchmarks in the W4A4 (per-tensor) setting, achieving gains that exceed 10~ppt J\&F for VOS, and 7.7~ppt mIoU for image segmentation. For the per-channel configurations, LSQ+, and LSQ~\cite{esser2019lsq} fail to converge, consistent with the limitations previously reported in~\cite{li2022mqbenchreproducibledeployablemodel}; results are thus omitted. The training curves for LSQ+ are represented in Figure~\ref{fig:lsq_loss} and show that LSQ+ begins with a large performance gap compared to Q-SAM2; while the gap narrows slightly over time, LSQ+ still converges to significantly lower accuracy. This suggests that LSQ+ may require more training data or epochs to achieve competitive results, reinforcing the efficiency and stability of the Q-SAM2 approach.

\subsubsection{Comparison with PTQ methods}
\label{sec:training_ptq}
\begin{table}[htbp]

    \centering
    \tiny

    \caption{Comparison between Q-SAM2 and PTQ methods based on the original SAM~\cite{kirillov2023segany}. }

    \label{tab:comparison_ptq}
    \resizebox{0.48\textwidth}{!}{
    \begin{tabular}{ccc|c|c|c}

        \toprule

        \textbf{Encoder} & \textbf{Precision} & \textbf{Size W (MB)} &\textbf{Method} &  \textbf{mAP} & \textbf{FP}  \\
        \midrule
        \multirow{6}{*}[-0.5mm]{ViT-H} & \multirow{3}{*}{W6A6} & \multirow{3}{*}{477} & PTQ4SAM~\cite{PTQ4SAMLv} & 48.7 &  \multirow{6}{*}[-0.5mm]{49.1}\\
        &&&  QDROP~\cite{wei2023qdroprandomlydroppingquantization} & 48.3 \\
        &&&  BRECQ~\cite{li2021brecq} & 46.0  \\
        \addlinespace[0.3em]
        \cline{2-5} 
        \addlinespace[0.3em]
        & \multirow{3}{*}{W4A4} & \multirow{3}{*}{317} & PTQ4SAM~\cite{PTQ4SAMLv} & 43.9 \\
        &&&  QDROP~\cite{wei2023qdroprandomlydroppingquantization} & 41.7 \\
        &&&  BRECQ~\cite{li2021brecq} & 17.6  \\
        \midrule
        \multirow{6}{*}[-0.5mm]{ViT-L} & \multirow{3}{*}{W6A6} & \multirow{3}{*}{230} & PTQ4SAM~\cite{PTQ4SAMLv} & 48.3 & \multirow{6}{*}[-0.5mm]{48.6}\\
        &&&  QDROP~\cite{wei2023qdroprandomlydroppingquantization} & 47.5 \\
        &&&  BRECQ~\cite{li2021brecq} & 46.6  \\
        \addlinespace[0.3em]
        \cline{2-5} 
        \addlinespace[0.3em]
        & \multirow{3}{*}{W4A4} & \multirow{3}{*}{154} & PTQ4SAM~\cite{PTQ4SAMLv} & 36.6 \\
        &&&  QDROP~\cite{wei2023qdroprandomlydroppingquantization} & 27.5 \\
        &&&  BRECQ~\cite{li2021brecq} & 12.3  \\
        \midrule
        \multirow{6}{*}[-0.5mm]{ViT-B} & \multirow{3}{*}{W6A6} & \multirow{3}{*}{71} & PTQ4SAM~\cite{PTQ4SAMLv} & 40.4 & \multirow{6}{*}[-0.5mm]{44.5}\\
        &&&  QDROP~\cite{wei2023qdroprandomlydroppingquantization} & 38.9 \\
        &&&  BRECQ~\cite{li2021brecq} & 31.8  \\
        \addlinespace[0.3em]
        \cline{2-5} 
        \addlinespace[0.3em]
        & \multirow{3}{*}{W4A4} & \multirow{3}{*}{47} & PTQ4SAM~\cite{PTQ4SAMLv} & 14.4 \\
        &&&  QDROP~\cite{wei2023qdroprandomlydroppingquantization} & 11.2 \\
        &&&  BRECQ~\cite{li2021brecq} & 3.6  \\

        \midrule
         &\cellcolor[HTML]{F3F3F3} W2A4 & \cellcolor[HTML]{F3F3F3}38.8 & &\cellcolor[HTML]{F3F3F3}45.8 & \multirow{2}{*}{49.6}\\ 
        \multirow{-2}{*}{Hiera-B+} &\cellcolor[HTML]{F3F3F3} W2A2 & \cellcolor[HTML]{F3F3F3}38.8 & & \cellcolor[HTML]{F3F3F3}39.2 \\
        
    
         \addlinespace[0.3em]
        \cline{2-3}   \cline{5-6} 
        \addlinespace[0.3em]
    
         & \cellcolor[HTML]{F3F3F3}W2A4 & \cellcolor[HTML]{F3F3F3}32.0 &&\cellcolor[HTML]{F3F3F3}44.3 & \multirow{2}{*}{48.4}\\
        \multirow{-2}{*}{Hiera-S} &\cellcolor[HTML]{F3F3F3} W2A2 &\cellcolor[HTML]{F3F3F3}32.0 && \cellcolor[HTML]{F3F3F3}36.1 \\
        
        \addlinespace[0.3em]
        \cline{2-3}   \cline{5-6} 
        \addlinespace[0.3em]
    
         & \cellcolor[HTML]{F3F3F3}W2A4 & \cellcolor[HTML]{F3F3F3}30.2 & &\cellcolor[HTML]{F3F3F3}43.8 & \multirow{2}{*}{48.0}\\
        \multirow{-2}{*}{Hiera-T} & \cellcolor[HTML]{F3F3F3}W2A2 &\cellcolor[HTML]{F3F3F3}30.2 & \multirow{-6}{*}[0.8mm]{\textbf{Q-SAM2}} &\cellcolor[HTML]{F3F3F3}35.4 \\

        \bottomrule
    \end{tabular}
    }
\end{table} 
We benchmark our QAT solution, Q-SAM2, against relevant PTQ quantization baselines. Our comparison includes PTQ4SAM~\cite{PTQ4SAMLv}, an algorithm specifically developed for the original SAM architecture~\cite{kirillov2023segany}. We also report two leading general PTQ schemes, QDROP~\cite{wei2023qdroprandomlydroppingquantization} and BRECQ~\cite{li2021brecq}, known for their strong general performance. Table~\ref{tab:comparison_ptq} summarizes the superior accuracy-compression trade-off achieved by Q-SAM2. We show that our method pushes state-of-the-art performance into the ultra-low bit regime, offering significant size reductions without sacrificing fidelity. Q-SAM2 achieves exceptional compression, with the Q-SAM2-B+ model requiring only 38.8 MB for storing the weights. This results in a $\sim 8.1\text{x}$ reduction in model size compared to the largest baseline, PTQ4SAM-H ($\text{W4A4}, 317 \text{ MB}$). Our smallest Q-SAM2-S and Q-SAM2-T configurations further reduce size to approximately $30\text{ MB}$. Crucially, the Q-SAM2 models retain accuracy far beyond similarly compressed or even larger baselines. The Q-SAM2-B+ ($\text{W2A4}, 45.8 \text{ mAP}$) is competitive with the largest PTQ4SAM-H ($\text{W4A4}, 43.9 \text{ mAP}$). Furthermore, Q-SAM2-B+ remains robust even in the most aggressive W2A2 setting, retaining $39.2 \text{ mAP}$, which surpasses the $\text{W4A4}$ performance of the larger PTQ4SAM-L ($\text{W4A4}, 36.6 \text{ mAP}$).

\subsection{SAM2 Weight and Activation Distributions}
\label{sec:distributions}
\begin{figure}[t]
    \centering
    \includegraphics[width=\linewidth]{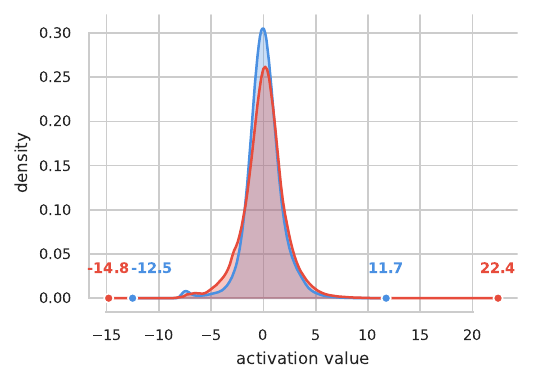}
    \caption{Comparison of the kernel density estimates across 50 random input generating the lowest (blue) and highest (red) total magnitude response in the \textit{image\_encoder.trunk.blocks.1.mlp.layers.0} layer of the B+ model. Markers indicate the precise minimum and maximum values, highlighting the shift in dynamic range between distinct inputs.}
    \label{fig:activations_dist}
\end{figure}
As discussed in Section~\ref{sec:challenges}, the core challenge in quantizing SAM2 stems from extreme weight and activation distributions. This appendix provides the empirical foundation for that assertion and justifies the introduction of our VRC and LSC methods.

We quantify the severity of weight outliers in the B+ encoder by analyzing the scale difference between the weight distribution's core and its extremes. Specifically, we examine the ratio of the standard deviation ($\sigma$) to the absolute maximum value ($\text{max}(|W|)$). We found that on average the absolute maximum value is located at \(13.6\sigma\)  with a peak of \(39.0\sigma\). These results confirm that high-magnitude outliers reside far from the central mass of the distribution, consuming a disproportionate share of the quantization range.

Furthermore, we demonstrate that the quantization challenge extends significantly to activations. Beyond exhibiting large value ranges, we observe substantial distribution shifts within the same layer during inference. This inherent statistical instability, which we visualize in Figure~\ref{fig:activations_dist}, underscores the critical need for a stabilization mechanism like momentum-stabilized clipping.
\begin{table}[htbp]

    \centering
    \tiny

    \caption{Ablations of value \(\alpha\) from Equation~\ref{eq:alpha_clipping} using model B+ and W2A4 configuration.}

    \label{tab:alpha_minmax}
    \resizebox{0.48\textwidth}{!}{
    \begin{tabular}{cccc}

        \toprule
        \textbf{Dataset} & \textbf{Clipping ($\alpha$)} & \textbf{VRC}  & \textbf{mIoU} \\
        \midrule
        \multirow{3}{*}{COCO2017~\cite{cocodataset}} &\myxmark &\myxmark & 39.0\\
        & \cellcolor[HTML]{F3F3F3}2.6 &\cellcolor[HTML]{F3F3F3}\myxmark & \cellcolor[HTML]{F3F3F3}43.3 \\
        & \cellcolor[HTML]{F3F3F3}2.6 &\cellcolor[HTML]{F3F3F3}\mycheckmark & \cellcolor[HTML]{F3F3F3}44.7 \\
        \bottomrule
    \end{tabular}
    }
\end{table} 
Building upon these observations, we investigate fixed clipping for weights and activations as an initial outlier mitigation strategy. This involves truncating values based on statistical bounds, specifically using the rule 
\begin{equation}
\mu \pm \alpha \sigma,
\label{eq:alpha_clipping}
\end{equation}
where $\mu$ and $\sigma$ are the distribution mean and standard deviation, and $\alpha$ is a fixed scaling factor. We apply Equation~\ref{eq:alpha_clipping} on a MinMax approach where we calculate the statistics per-tensor for activations and per-channel for weights, then applying a fixed \(\alpha=2.6\). Table~\ref{tab:alpha_minmax} shows the results of this experiment, and confirms the effectiveness of statistical clipping, with a 4.3 ppt improvement in accuracy over the unclipped baseline. Furthermore, our VRC enhances the performance of the MinMax approach, demonstrating additional 1.4ppt gains beyond those achieved by simple static bounds.

\subsection{Qualitative Results}
\label{sec:qres}
\begin{figure*}[htb]
  \centering
  \includegraphics[width=\textwidth]{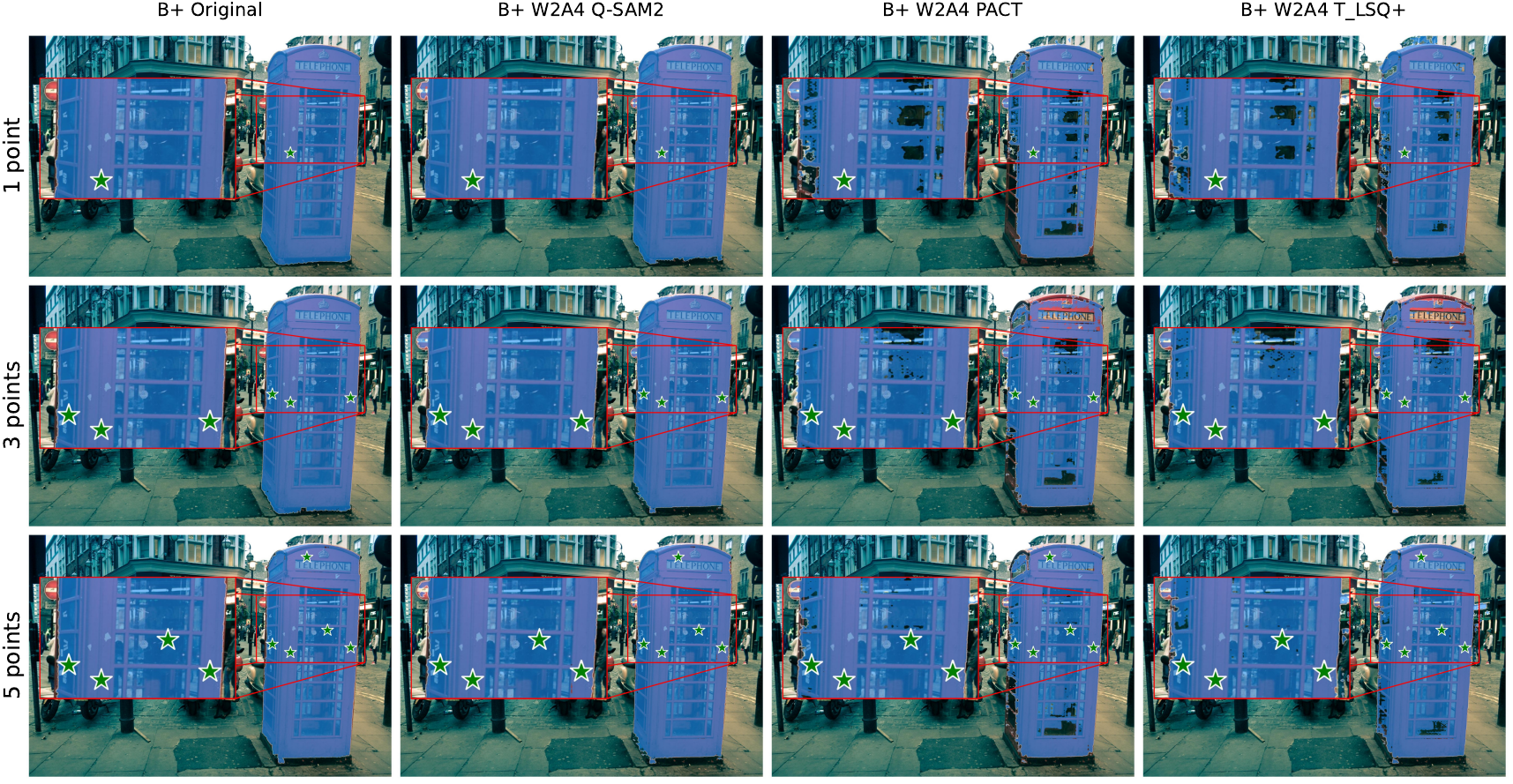} 
  \caption{Qualitative results of SAM2, QSAM2(ours), PACT, and T\_LSQ+ on the base plus architecture. From top to bottom, we add 1, 3, and 5 prompt points.}
  \label{fig:qres_prog_0}
\end{figure*}

\begin{figure*}[htb]
  \centering
  \includegraphics[width=\textwidth]{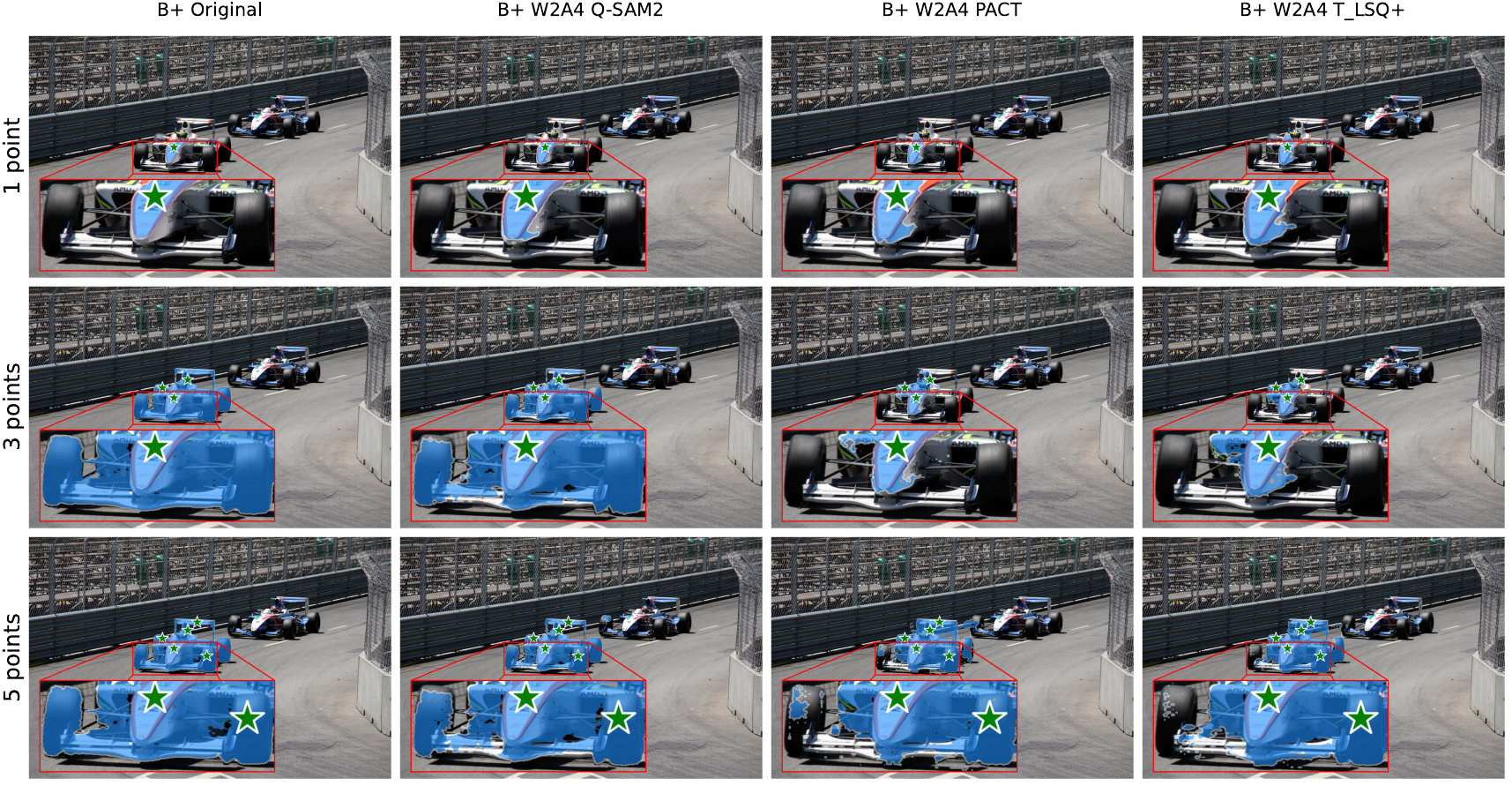} 
  \caption{Qualitative results of SAM2, QSAM2(ours), PACT, and T\_LSQ+ on the base plus architecture. From top to bottom, we add 1, 3, and 5 prompt points.}
  \label{fig:qres_prog_2}
\end{figure*}

\begin{figure*}[htb]
  \centering
  \includegraphics[width=\textwidth]{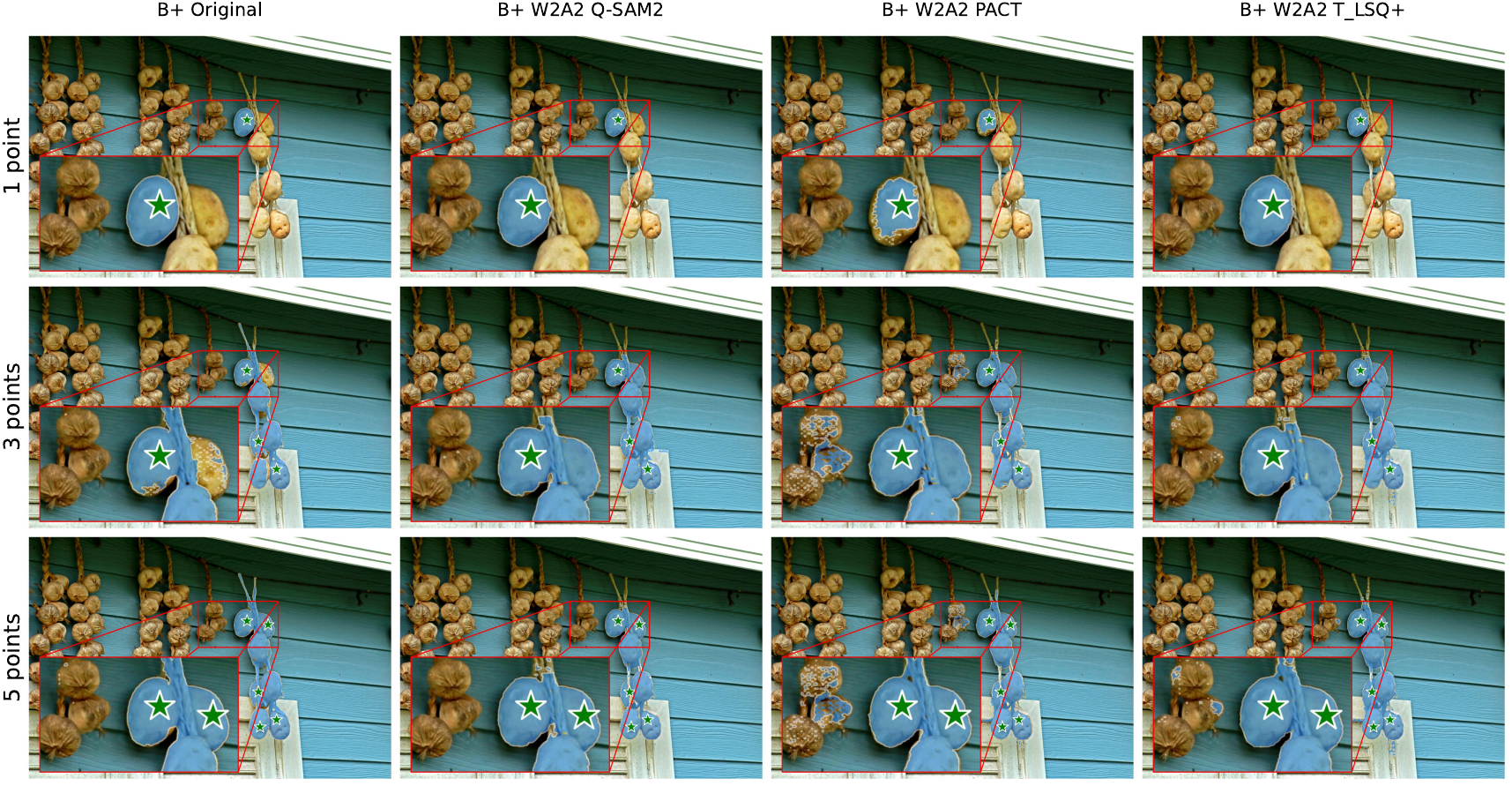} 
  \caption{Qualitative results of SAM2, QSAM2(ours), PACT, and T\_LSQ+ on the base plus architecture. From top to bottom, we add 1, 3, and 5 prompt points.}
  \label{fig:qres_prog_3}
\end{figure*}

We have extensively benchmarked the original variants of SAM2, the MinMax baseline, and our Q-SAM2 algorithm. The variation includes different architectures, prompts, and quantization configurations for a fair comparison in real-world scenarios. Figure~\ref{fig:qres_prog_0}, Figure~\ref{fig:qres_prog_2}, and Figure~\ref{fig:qres_prog_3} show the impact of adding more input points as prompt. We observe that our algorithm provides cleaner masks compared to PACT, and T\_LSQ+ algorithms with the same quantization scheme. We also note that by providing one prompt point only, typically all variants of the algorithms tend to provide a clean mask. However, and that is a well-studied behavior of the original SAM model, this mask might be only a part of the user's intended selection, such as the front of the racing car as in Figure~\ref{fig:qres_prog_2}, or a single garlic bulb as in Figure~\ref{fig:qres_prog_3}. Henceforth, interesting examples include those that require multiple prompts to select the entire object. Those 3-point or 5-point prompts trigger differences among the quantized models. Compared against PACT, and T\_LSQ+, our 2-bit models require fewer prompt points to recover a larger fraction of the entire object, producing a cleaner and better result.

\begin{figure*}[htb]
  \centering
  \includegraphics[width=\textwidth]{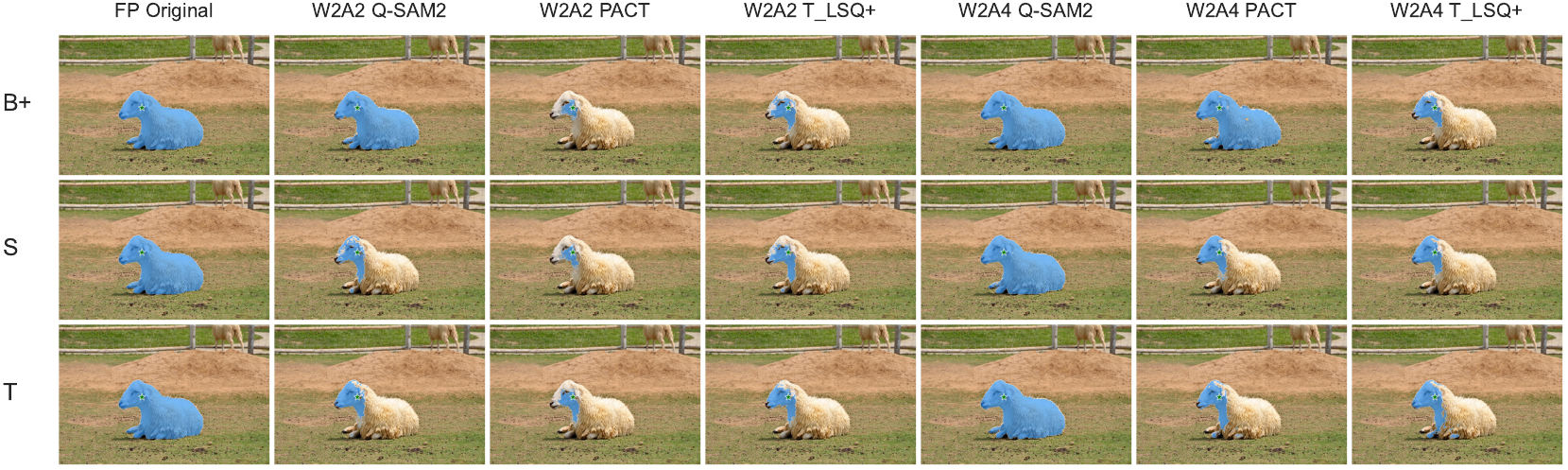} 
  \caption{Qualitative results of variants of models with 1 point as prompt.}
  \label{fig:qres_all_sheep1}
\end{figure*}

\begin{figure*}[htb]
  \centering
  \includegraphics[width=\textwidth]{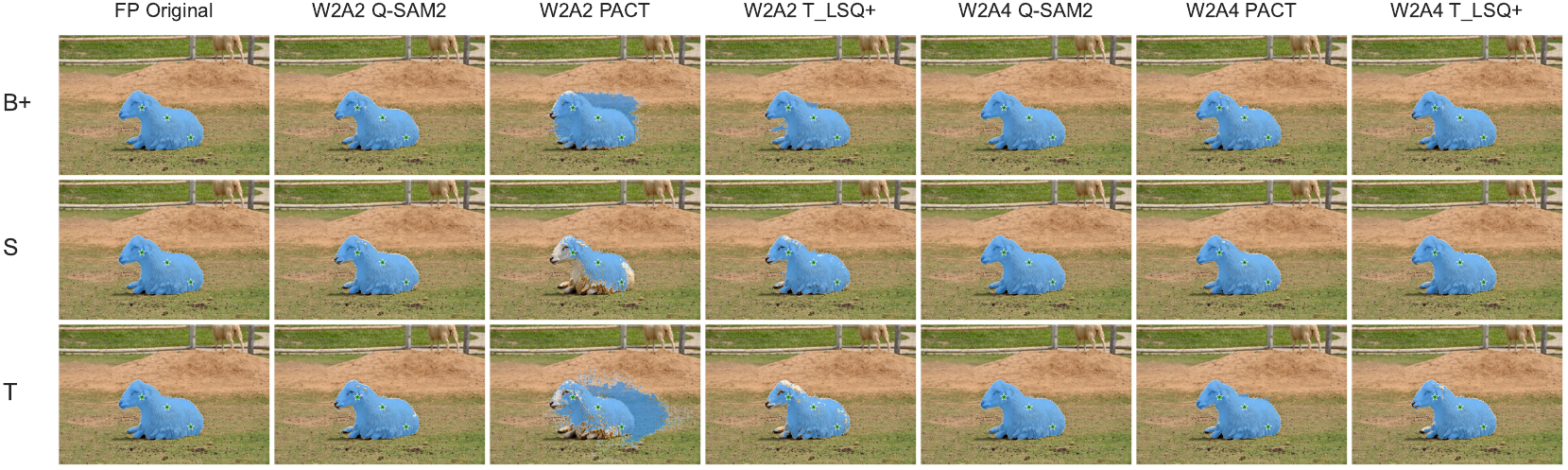} 
  \caption{Qualitative results of variants of models with 3 points as prompt.}
  \label{fig:qres_all_sheep3}
\end{figure*}

Figure~\ref{fig:qres_all_sheep1}, Figure~\ref{fig:qres_all_sheep3} show the entire family of models including SAM2 base plus, small, and tiny architectures at full precision, as well as all variants of PACT, T\_LSQ+, and our Q-SAM2 models with 2-bit quantization for weights and 2 and 4 bits for activations. We first analyze the visual results prompted by a single point (Image~\ref{fig:qres_all_sheep1}). In this challenging scenario, many baseline models erroneously segment only the object's head. Our Q-SAM2 family (W2A4 variants) and the B+ W2A2 model yield segmentation maps comparable to the full-precision network, whereas other baselines struggle significantly. We examine the second image (prompted by three points), where nearly all models correctly identify the target object. However, a closer inspection reveals distinct qualitative differences between our proposed Q-SAM2 solution and the competing baselines.

Finally, Figures~\ref{fig:qres_video1} and~\ref{fig:qres_video2} present a qualitative comparison on a video from the SA-V test set for the semi-supervised Visual Object Segmentation (VOS) task. The models included in this comparison are SAM2.1, the baselines, and our proposed Q-SAM2 for the W2A2 schema. The first column of Figure~\ref{fig:qres_video1} shows the initial ground-truth mask used to prompt the models, as required in the semi-supervised VOS setup. Although T\_LSQ+ generally succeeds in segmenting the target object, Q-SAM2 produces cleaner masks with fewer border artifacts compared to the baselines.

\begin{figure*}[htb]
  \centering
  \includegraphics[width=\textwidth]{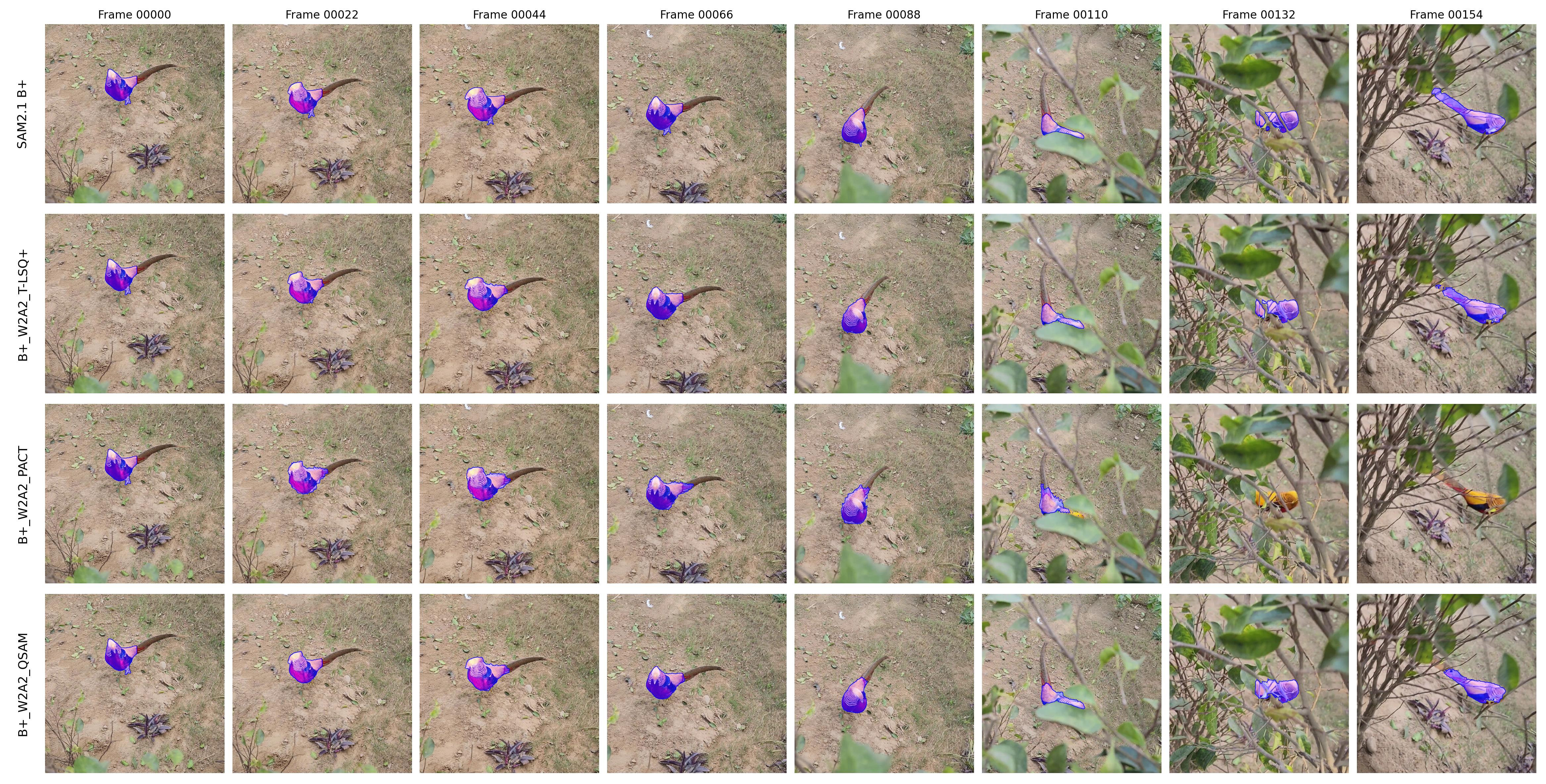} 
  \caption{Qualitative results of variants of models SAM2, MinMax, and QSAM2(ours) on the base plus architecture for a video sequence. The first column shows the input mask provided in the first frame, while the remaining columns display the predicted segmentations for each model across subsequent frames.}
  \label{fig:qres_video1}
\end{figure*}
\begin{figure*}[htb]
  \centering
  \includegraphics[width=\textwidth]{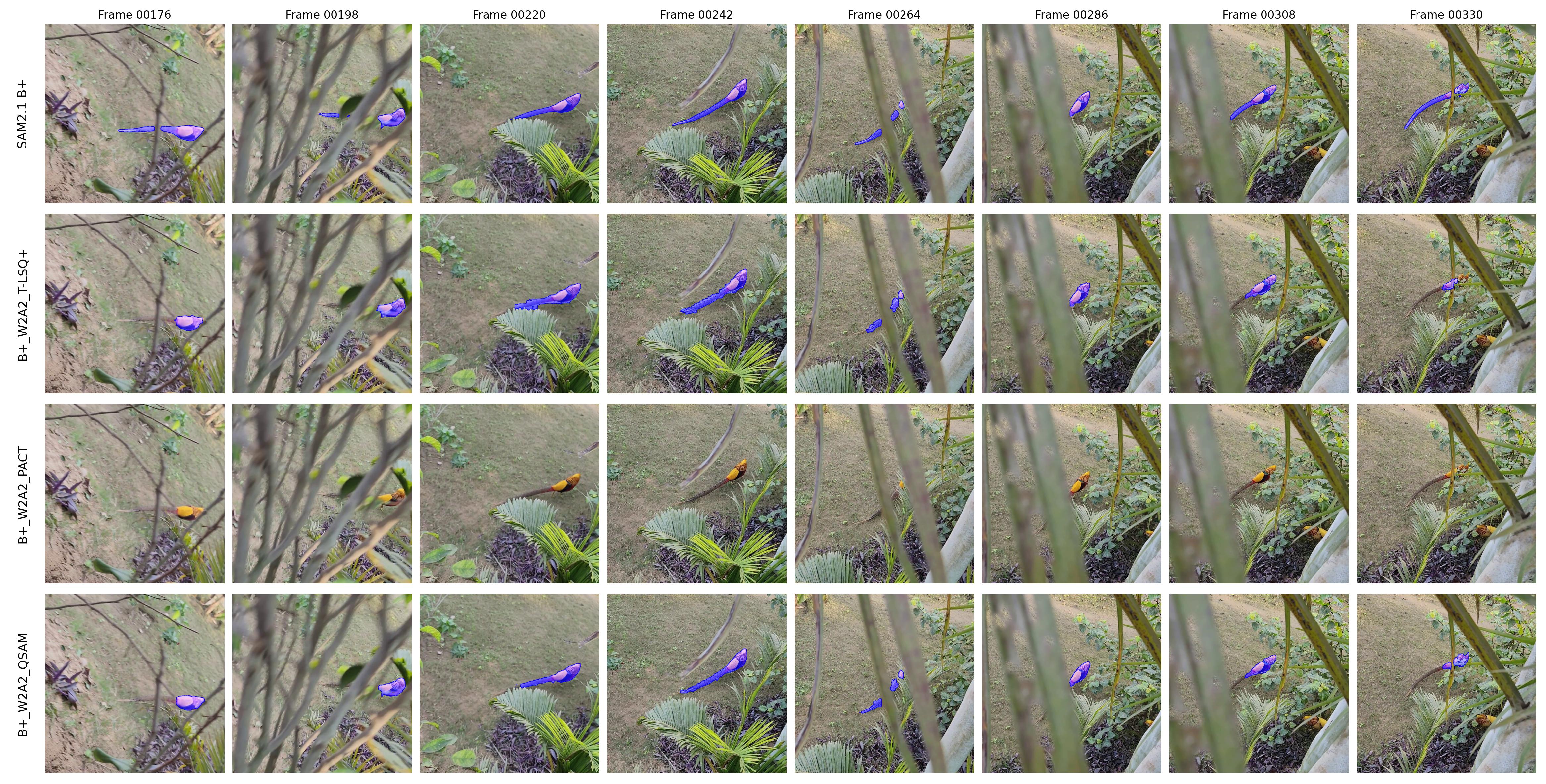} 
  \caption{Continuation of qualitative results from Figure~\ref{fig:qres_video1}, showing additional frames from the same video sequence.}
  \label{fig:qres_video2}
\end{figure*}

\end{document}